\documentclass[10pt,journal,compsoc]{IEEEtran}
\ifCLASSOPTIONcompsoc
  \usepackage[nocompress]{cite}
\else
  \usepackage{cite}
\fi
\ifCLASSINFOpdf
\else
\fi
\usepackage{amsmath}
\usepackage{amsmath}
\usepackage{amsfonts}
\usepackage{bm}%加粗
\usepackage{mathrsfs}
\usepackage[nocompress]{cite}

\usepackage[margin=0pt,font=small,labelfont=bf,labelsep=endash,tableposition=top]{caption}
\usepackage{amsmath}
\usepackage{array}

\usepackage[caption=false,font=footnotesize,labelfont=sf,textfont=sf]{subfig}
\usepackage{fixltx2e}

\usepackage{url}
\usepackage{graphicx}
\usepackage{amsmath}
\usepackage{amssymb}
\usepackage{booktabs}
\usepackage{epsfig}
\usepackage{epstopdf}
\usepackage{makecell,multirow,diagbox}
\usepackage{color}
\usepackage{soul}
\usepackage{url}
\usepackage{amsmath}
\usepackage{array}

\usepackage[utf8]{inputenc}
\usepackage{xcolor}
\usepackage{hyperref}
\usepackage{helvet}
\usepackage{courier}
\usepackage{bm}
\usepackage{bbm}
\usepackage{wrapfig}
\usepackage{picinpar}
\usepackage{comment}

\usepackage{cite}
\usepackage{balance}

\usepackage{color}
\newcommand{\wu}[1]{\textcolor[rgb]{1,0,0} {#1}}

\newcommand{\pei}[1]{\textcolor[rgb]{0,1,0} {#1}}

\begin{document}
%\title{Single Shot Self-Reliant Scene Text Spotter by Decoupling Recognition from Detection}
\title{Single Shot Self-Reliant Scene Text Spotter by Decoupled yet Collaborative Detection and Recognition}

\author{Jingjing Wu$^*$,
        Pengyuan Lyu$^*$,
        Guangming Lu,~\IEEEmembership{Member,~IEEE},
        Chengquan Zhang, 
        and~Wenjie Pei\textsuperscript{\dag}

\thanks{$^*$Equal contribution.}
\thanks{\textsuperscript{\dag}Wenjie Pei is the corresponding author.}
\IEEEcompsocitemizethanks{\IEEEcompsocthanksitem Jingjing Wu, Guangming Lu and Wenjie Pei are with the Department of
Computer Science, Harbin Institute of Technology at Shenzhen, Shenzhen
518057, China (E-mails: jingjingwu\_hit@outlook.com; luguangm@hit.edu.cn; wenjiecoder@outlook.com).\protect\\
\IEEEcompsocthanksitem Pengyuan Lyu and Chengquan Zhang  are with Department  of  Computer Vision 
Technology, Baidu  Inc (E-mails: lvpyuan@gmail.com; zhangchengquan@baidu.com).}%; yaokun01@baidu.com).}% <-this % stops an unwanted space
%\thanks{Manuscript received xx, 20xx; revised xx, 20xx.}
}

\IEEEtitleabstractindextext{%
\begin{abstract}
Typical text spotters follow the two-stage spotting paradigm which detects the boundary for a text instance first and then performs text recognition within the detected regions. Despite the remarkable progress of such spotting paradigm, an important limitation is that the performance of text recognition depends heavily on the precision of text detection, resulting in the potential error propagation from detection to recognition.  In this work, we propose the single shot Self-Reliant Scene Text Spotter v2 (\textbf{SRSTS v2}), which circumvents this limitation by decoupling recognition from detection while optimizing two tasks collaboratively. Specifically, our \textbf{SRSTS v2} samples representative feature points around each potential text instance, and conducts both text detection and recognition in parallel guided by these sampled points. Thus, the text recognition is no longer dependent on detection, thereby alleviating the error propagation from detection to recognition. Moreover, the sampling module is learned under the supervision from both detection and recognition, which allows for the collaborative optimization and mutual enhancement between two tasks. Benefiting from such sampling-driven concurrent spotting framework, our approach is able to recognize the text instances correctly even if the precise text boundaries are challenging to detect. Extensive experiments on four benchmarks demonstrate that our method compares favorably to state-of-the-art spotters.

\end{abstract}

}

\maketitle

\IEEEdisplaynontitleabstractindextext

% \fi
\IEEEpeerreviewmaketitle

\IEEEraisesectionheading{\section{Introduction}\label{sec:introduction}}
% 1. two-stage spotters and their limitations
\IEEEPARstart{S}{cene} text spotting, which aims to detect and recognize text from natural images simultaneously, has extensive applications ranging from intelligent transportation, document reading to vision search based on texts. Despite the great success achieved in recent years, it remains a challenging task due to diverse text appearances in terms of shape, size, and style.
\begin{figure}[t]
  \centering
  %  \vspace{-10pt} 
  \includegraphics[width=\linewidth]{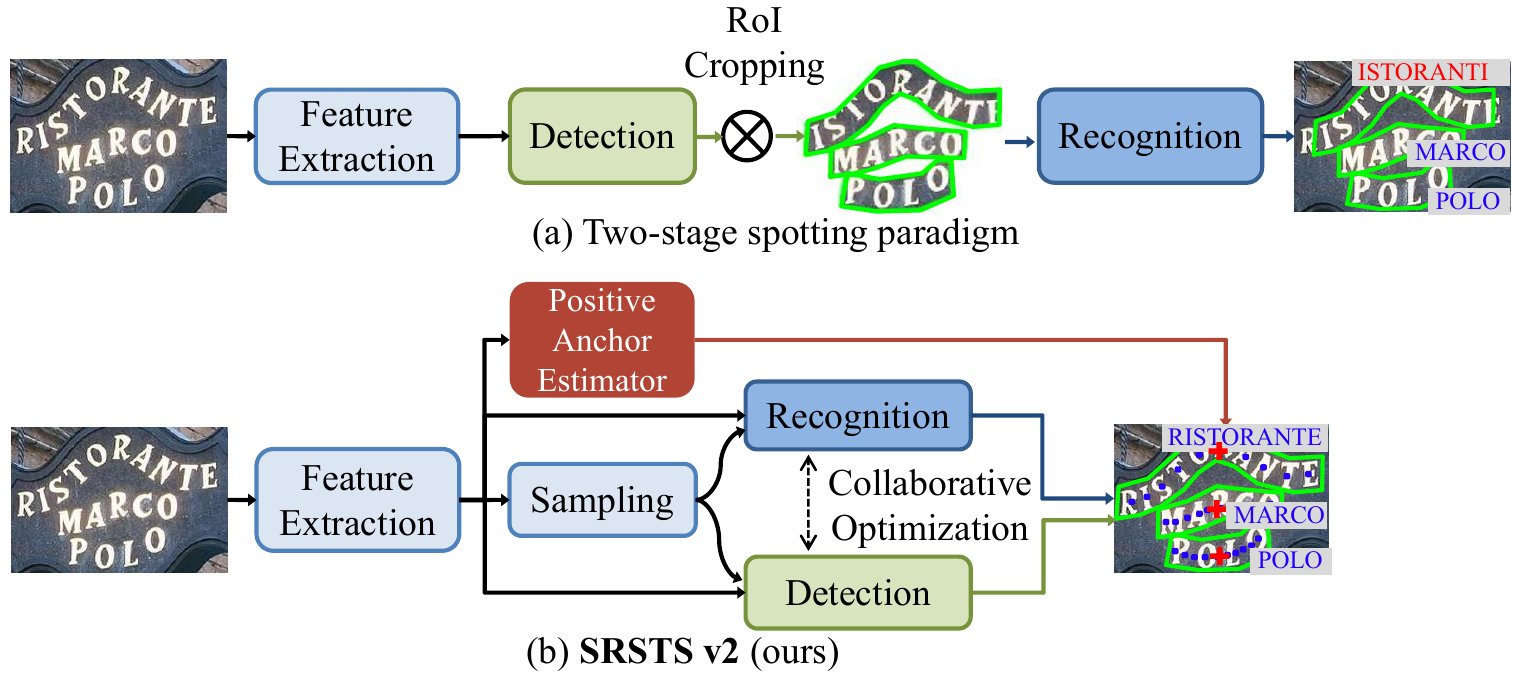}
%   \includegraphics[width=\linewidth]{figs/intro1.pdf}
    % \vspace{-20pt}   
  \caption{Unlike typical two-stage spotting paradigm (a) that recognizes text based on the detected and cropped RoIs, our \textbf{SRSTS v2} (b) estimates a positive anchor point (‘$+$’s in \textcolor[RGB]{255,0,0}{red}) for each potential text instance, and samples representative feature points (\textcolor[RGB]{0,0,255}{blue} points) around anchor points by a specially designed sampling module. As a result, our \textbf{SRSTS v2} is able to conduct text detection and recognition in parallel guided by the sampled points, thereby eliminating the dependence of recognition on the detection results. Furthermore, the sampling module is learned by the joint supervision from both the detection and recognition tasks, which enables the collaborative optimization and mutual enhancement between two tasks.} %Instead of recognizing text from the detected and cropped RoIs as previous two-stage methods, the detection module and recognition module in our \textbf{SRSTS} are conducted in parallel  guided by positive anchor points predicted by anchor estimation module. As a result, our recognition module can recognize texts correctly even though the detection is not precise. decoupled and bridged by positive anchor points predicted by anchor estimation module. As a result, our recognition module can recognize texts correctly even though the detection is not precise. 
  %}
   \label{fig:teaser}
%  \vspace{-8pt} 
\end{figure}

Most existing methods for text spotting~\cite{lyu2018mask,liao2019mask,liao2020mask,qin2019towards,feng2019textdragon,wang2020all,liu2020abcnet,liu2021abcnet} follow the typical two-stage spotting paradigm, which first performs text detection to locate the position of text instances in the first stage and then recognizes the text within the detected region of interests (RoIs) in the second stage. While remarkable progress has been made by these methods, there are two potential limitations preventing those methods from achieving better performance. One is that the performance of text recognition relies heavily on the precision of text detection. As a result, such methods suffer from the error propagation from text detection to recognition. The other limitation is that the RoI cropping operation does not filter the noise in the background. Besides, it also brings about information loss caused by the operation of pooling or interpolation on feature maps, which would also adversely affect the performance of text recognition. Considering the instance in Figure ~\ref{fig:teaser}, the curved text `ristorante' is extremely difficult to detect precisely and its irregular shape is also not so friendly to RoI cropping. Thus, the two-stage text spotters usually fail to locate the accurate boundary during text detection. In addition, the RoI cropping operation further introduces some errors, especially near the boundaries of the text. Consequently, the characters `r' and `e' are missed or wrongly recognized in the recognition stage, leading to erroneous end-to-end spotting results.

Recently, some single shot text spotters ~\cite{xing2019convolutional,qiao2020mango, wang2021pgnet,zhang2022text,kittenplon2022towards} integrate both the text detection and recognition into one-stage. Benefiting from such one-stage framework, the RoI cropping operation is not necessary, which thereby eliminates the aforementioned disadvantages of RoI cropping. Nevertheless, these methods perform recognition depending on accurate detection of text region~\cite{xing2019convolutional} or center line of text instances~\cite{wang2021pgnet}, thus they still have the limitation of error propagation from detection to recognition. Qiao \textit{et al.}~\cite{qiao2020mango} proposed MANGO which treats the text spotting  task as a sole text recognition problem. However, extra character-level annotation for supervision is required, which is extremely label-consuming.

In this work, we propose the single shot Self-Reliant Scene Text Spotter v2 (\textbf{SRSTS v2}), which decouples recognition from detection and thereby reduces the dependence of recognition on the detection results. Inspired by the concept of ‘anchor’ in ~\cite{ren2015faster}, we view each pixel of feature maps as an `anchor point' for text spotting to provide location guidance for text detection as well as recognition. To be specific, the proposed \textbf{SRSTS v2} estimates a positive anchor point for each potential text instance, as shown in Figure~\ref{fig:teaser}. Meanwhile, a specialized sampling module is designed to sample representative feature points for each anchor point. The sampled feature points around each positive anchor point are leveraged to perform both text detection and recognition in parallel. Thus, the sampling module plays two important roles. 1) It decouples the task of text recognition from detection. Both the detection and recognition are conducted based on the sampled feature points in parallel, thus the task of text recognition is no longer dependent on the detection results and thus the precise text detection is not essential for recognition. 2) The sampling module enables collaborative optimization between detection and recognition. Both the supervision from detection and recognition tasks guide the learning of the sampling module, the optimized sampling module in turn enhances the performance of both tasks. %allowing for the collaborative optimization between two tasks. 

The proposed sampling-driven concurrent spotting framework yields three merits of our \textbf{SRSTS v2}. First, the performance of recognition is not strictly limited by the precision of detection, which is particularly advantageous in challenging scenarios for detection since estimating a rough positive anchor point and sampling representative points for a text instance is quite easier than predicting its precise boundary. Second, the detection and recognition can be optimized collaboratively to potentially enhance each other. Third, since the recognition of \textbf{SRSTS v2} does not rely on precise detection results, it is feasible to train \textbf{SRSTS v2} with limited supervision for detection. Such learning paradigm makes it possible to reduce the annotation cost for text detection substantially than the typical spotting methods, let alone the character-level annotations required by the previous single shot spotters ~\cite{xing2019convolutional, qiao2020mango}.
To conclude, the main contributions of this work can be summarized as follows: 
\begin{itemize}[]
    % \item We propose to perform text spotting by adaptively sampling informative points within the text instance. Benefiting from the designed effective spotting mechanism, \textbf{SRSTS v2} is able to recognize the text instances correctly even though the text boundaries are challenging to detect.
    % \item An interaction mechanism is introduced to keep information interaction between the decoupled detection and recognition.
    % \item We explore a series of efficient feature enhancement components to improve the spotting performance of proposed method, including deformable transformer encoder and \wu{self-attention layers}.
    % \item As the recognition is free from precise detection results, our proposed \textbf{SRSTS v2} can be trained in a weakly-supervised manner. 
    % \item  We conduct extensive experiments to evaluate our method, including various ablation studies to investigate the effectiveness of proposed components and the comprehensive comparisons to the state-of-the-art methods for text spotting on both regular-shaped benchmarks and arbitrary-shaped benchmarks. All these experiments validate the advantages of our method. 
    \item Our proposed \textbf{SRSTS v2} conducts both text detection and recognition in parallel based on the sampled representative feature points by the specially designed sampling module. Such design not only eliminates the dependence of text recognition on the detection results, thereby circumventing the potential error propagation from detection to recognition, but also allows for the collaborative optimization between two tasks to potentially enhance each other.
    \item Benefiting from the designed effective spotting mechanism, \textbf{SRSTS v2} is able to recognize the text instances correctly even though the text boundaries are challenging to detect.
    \item The decoupling of text recognition from detection makes it feasible for our proposed \textbf{SRSTS v2} to be trained with limited annotation for text detection, which is a preliminary exploration of weakly supervised text spotting.
    \item We conduct extensive experiments to evaluate our method, including various ablation studies to investigate the effectiveness of core components of \textbf{SRSTS v2} and the comprehensive comparisons with the state-of-the-art methods for text spotting on both regular-shaped benchmarks and arbitrary-shaped benchmarks. All these experiments validate the advantages of our method.
\end{itemize}

This paper is extended from an earlier conference version~\cite{wu2022decoupling} which proposed the initial \textbf{SRSTS}. Compared with the prior version, this longer article is improved in three aspects. First, we redesign the detection module, which was previously designed based on YOLACT~\cite{bolya2019yolact} in the conference version~\cite{wu2022decoupling}. Our \textbf{SRSTS v2} performs both text detection and recognition based on the sampled representative points from the sampling module, yielding the sampling-driven concurrent spotting framework. Such newly proposed spotting framework enables the collaborative optimization between detection and recognition since both the supervision of two tasks contribute to the learning of the sampling module. The experiments in Section~\ref{sec: comparison} validates the distinct performance improvement in both detection and recognition by such collaborative optimization. Second, we apply the deformable transformer encoder~\cite{zhu2020deformable} to enhance the feature learning and employ the self-attention operation to capture the long-range dependency among the features of sampled points, which further boosts the overall performance by a large margin, as shown in Experiments in Section~\ref{sec: comparison}. Third, substantial additional experiments are conducted to evaluate our approach more comprehensively, including 1) thorough evaluation on the extended version \textbf{SRSTS v2}; 2) more detailed ablation study to obtain more insights into our approach; 3) full experiments on two more diverse datasets including Rotated ICDAR 2013 and CTW1500.

\section{Related Work}
%\wu{The previous methods usually treat scene text spotting as two separate sub-tasks : text detection and text recognition. In recent years, integrating text detection and recognition into an unified framework becomes a new trend. We will present brief introduction to existing methods which either focus on text detection or text recognition at first. The rest of this part will review the previous text spotting methods.}

%\subsection{Scene Text Detection}
%Scene text detection aims to locate text instance in the wild. Before the emergence of deep learning...

%\subsection{Scene Text Recognition}
%Scene text recognition aims to perform text decoding in cropped text image.
%The previous scene text spotters can be coarsely divided into two categories: two-stage scene text spotters and single shot scene text spotting methods.

%\subsection{Scene Text Spotting}
In recent years, scene text spotting has attracted extensive attention in the community of computer vision. The existing scene text spotters can be coarsely divided into two
categories: two-stage scene text spotters and single shot scene text
spotting methods.

\smallskip\noindent\textbf{Two-stage scene text spotting.} In the early stage, the scene text spotters ~\cite{jaderberg2016reading,liao2017textboxes,liao2018textboxes++} are always composed of separate text detector and recognizer. Jaderberg \textit{et al.} ~\cite{jaderberg2016reading} use a regression-based text detector and a word  classification  based recognizer to detect text and recognize text respectively. In ~\cite{liao2017textboxes, liao2018textboxes++}, a SSD ~\cite{liu2016ssd} based text detector and CRNN ~\cite{shi2016end} are employed to read text. Though substantial progress has been achieved, those methods still struggle to get better performance due to the sub-optimal between text detection and recognition.

To mitigate the sub-optimal between text detection and recognition, some end-to-end trainable frameworks are proposed.  Busta \textit{et al.} ~\cite{busta2017deep} first train text detector and recognizer separately and then fine-tune the two modules jointly. In \cite{li2017towards}, to learn text detector and recognizer in an end-to-end manner, a complex  curriculum learning paradigm is used. To train text detection and recognition jointly and steadily, some methods ~\cite{liu2018fots,wang2021pan++,liu2020abcnet,liu2021abcnet,qin2019towards} use the groundtruth boxes to extract features for text recognition. In ~\cite{lyu2018mask,liao2019mask, liao2020mask}, the above mentioned issue is relieved in the way of character segmentation. 

Detecting and recognizing text of arbitrary shapes has also attracted a lot of attention. Mask Textspotter  ~\cite{lyu2018mask} reads text by segmenting text regions and characters. Qin \textit{et al.} ~\cite{qin2019towards} proposed RoI masking which multiplies the cropped features with text instance segmentation masks. TextDragon ~\cite{feng2019textdragon} uses RoISlide to sample local features along the center line. Mask Textspotter v3 ~\cite{liao2020mask} proposed SPN (segmentation proposal network) to represent arbitrary-shape proposals. Text Perceptron ~\cite{qiao2020text} and Boundary ~\cite{wang2020all} represent the text boundary as a group of key points and apply the TPS (thin-platespline transformation) ~\cite{bookstein1989principal} to rectify irregular boundary. ABCNet~\cite{liu2020abcnet} and ABCNet v2 ~\cite{liu2021abcnet} innovatively use Bezier curve to fit curve contour and introduce BezierAlign operation to crop text feature map. 

%\wu{Typical end-to-end text spotters usually feed the recognition module with local features cropped from the shared feature map. Recently, some efforts have been made in feature extraction for recognition performance in end-to-end framework. GLASS ~\cite{ronen2022glass} proposes a fusion module to incorporate the global and local features to improve the robustness to scale. SwinTextspotter ~\cite{huang2022swintextspotter} exploits the feature interaction between detection and recognition. DLD ~\cite{chen2022dynamic} proposed a sequential knowledge distillation strategy and a dynamic resolution selector to find a suitable scale for recognition, achieving a better resolution performance balance. %how to deal with the performance degradation when faced with various scale GLASS ~\cite{ronen2022glass} propose to fuse global and local features.} 

Most of the aforementioned two-stage methods  conduct text detection and recognition serially and connect them by the RoI cropping operation. As a result, the recognition result is heavily affected by  detection and RoI cropping.

\smallskip\noindent\textbf{Single shot scene text spotting.} Recently several works 
attempt to integrate the detector and recognizer into a one-stage network to avoid the adverse effects of RoI cropping. CharNet ~\cite{xing2019convolutional} simultaneously predicts instance-level position information and character-level bounding box with character label. Based on the instance-level detection result, the text instance can be generated by grouping predicted characters. Recently PGNet ~\cite{wang2021pgnet} predicts various text region information in parallel and adopts point-gathering operation to gather the pixel-level character classification probability. To further reduce the dependence on detection module, MANGO ~\cite{qiao2020mango} treats the spotting task as a pure recognition task. 
% However, MANGO requires character-level annotations, and can't be utilized for real-time application. 
Inspired by recent objection detection methods such as DETR~\cite{carion2020end}, Deformable DETR~\cite{zhu2020deformable} and Pix2Seq ~\cite{chen2021pix2seq}, some works utilize the  encoder-decoder framework to locate and decode text instance. TTS ~\cite{kittenplon2022towards} and TESTR ~\cite{zhang2022text} both refine Deformable DETR to deal with text spotting task. SPTS ~\cite{peng2022spts} casts text spotting as a sequence modeling task and only uses single-point level text annotation for training.

Our \textbf{SRSTS v2} is also a single shot scene text spotter. To be specific, \textbf{SRSTS v2} estimates a positive anchor point for each potential text instance and performs text detection and text recognition in parallel based on sampled representative points by a specially designed sampling module, resulting in decoupling and collaborative optimization between detection and recognition. Our method outperforms the previous one-stage methods in the following advantages: 1) compared with ~\cite{wang2021pgnet}, the recognition of \textbf{SRSTS v2} does not rely on precise detection results; 2) compared with ~\cite{xing2019convolutional, qiao2020mango}, \textbf{SRSTS v2} only needs word-level annotations, while ~\cite{xing2019convolutional, qiao2020mango} require character-level annotations; 3) compared with ~\cite{kittenplon2022towards, zhang2022text, peng2022spts}, \textbf{SRSTS v2} achieves better spotting performance and is also able to be trained with limited annotation for text detection.%3) our method is able to run at real-time, while ~\cite{xing2019convolutional, qiao2020mango,zhang2022text, kittenplon2022towards} can not.}% }%1) compared to ~\cite{wang2021pgnet}, our method decouples recognition from detection, thus the recognition does not relay on precise detection results;  2) compared with ~\cite{xing2019convolutional, qiao2020mango}, ours method only needs word-level annotations, while ~\cite{xing2019convolutional, qiao2020mango} require character-level annotations; 3) our method is able to run at real-time, \wu{while ~\cite{xing2019convolutional, qiao2020mango,zhang2022text, kittenplon2022towards} can not.}

\section{Self-Reliant Scene Text Spotter}
Unlike typical scene text spotters that rely on text detection to acquire the precise boundaries of text instances for text recognition, the proposed \textbf{SRSTS v2} estimates a positive anchor point for each potential text instances, and samples representative feature points around each anchor point by specially designed Sampling Module. Consequently, it conducts text detection and recognition in parallel based on the sampled feature points. Thus, our \textbf{SRSTS v2} is able to decouple text recognition from text detection and eliminate the dependence of text recognition on the detection results, thereby circumventing the potential error propagation from detection to recognition. In particular, Sampling Module is learned under the supervision from both tasks of detection and recognition, which allows for the collaborative optimization and mutual enhancement between two tasks. We will first present the overall framework of the \textbf{SRSTS v2} and then describe how to estimate positive anchor points for text instances. Finally, we will elaborate on the text decoding mechanism of our approach, including the sampling strategy by Sampling Module, concurrent text detection and recognition, and collaborative optimization between these two tasks.%including decoupled text detection and recognition, and the information interaction between them.}

%The detection branch and recognition branch are both expected to guide the sampled points to cover more informative text positions. In our framework, text detection and recognition can be conducted in parallel, thereby circumventing the potential error propagation from detection to recognition and promoting each other to a certain extent. We will first present the overall framework of the \textbf{SRSTS v2}, and then describe how to estimate positive anchor points for text instances. Finally, we will elaborate on text detection and text recognition.  
%Unlike typical scene text spotters that rely on text detection to acquire the precise boundaries of text instances for text recognition, \lyu{the proposed Self-Reliant Scene Text Spotter (\textbf{SRSTS}) takes anchor points as reference and conducts text detection and recognition in parallel. As a result, our \textbf{SRSTS} is able to decouple text recognition from text detection and reduce the dependence of text recognition on the detection performance, thereby circumventing the potential error propagation from detection to recognition. We will first present the overall framework of the \textbf{SRSTS}, and then describe how to estimate positive anchor points for text instances. Finally we will elaborate on the text detection and text recognition.}
\subsection{Overall Framework}

\begin{figure*}[t]
   \centering
   \centerline{\includegraphics[width=\textwidth]{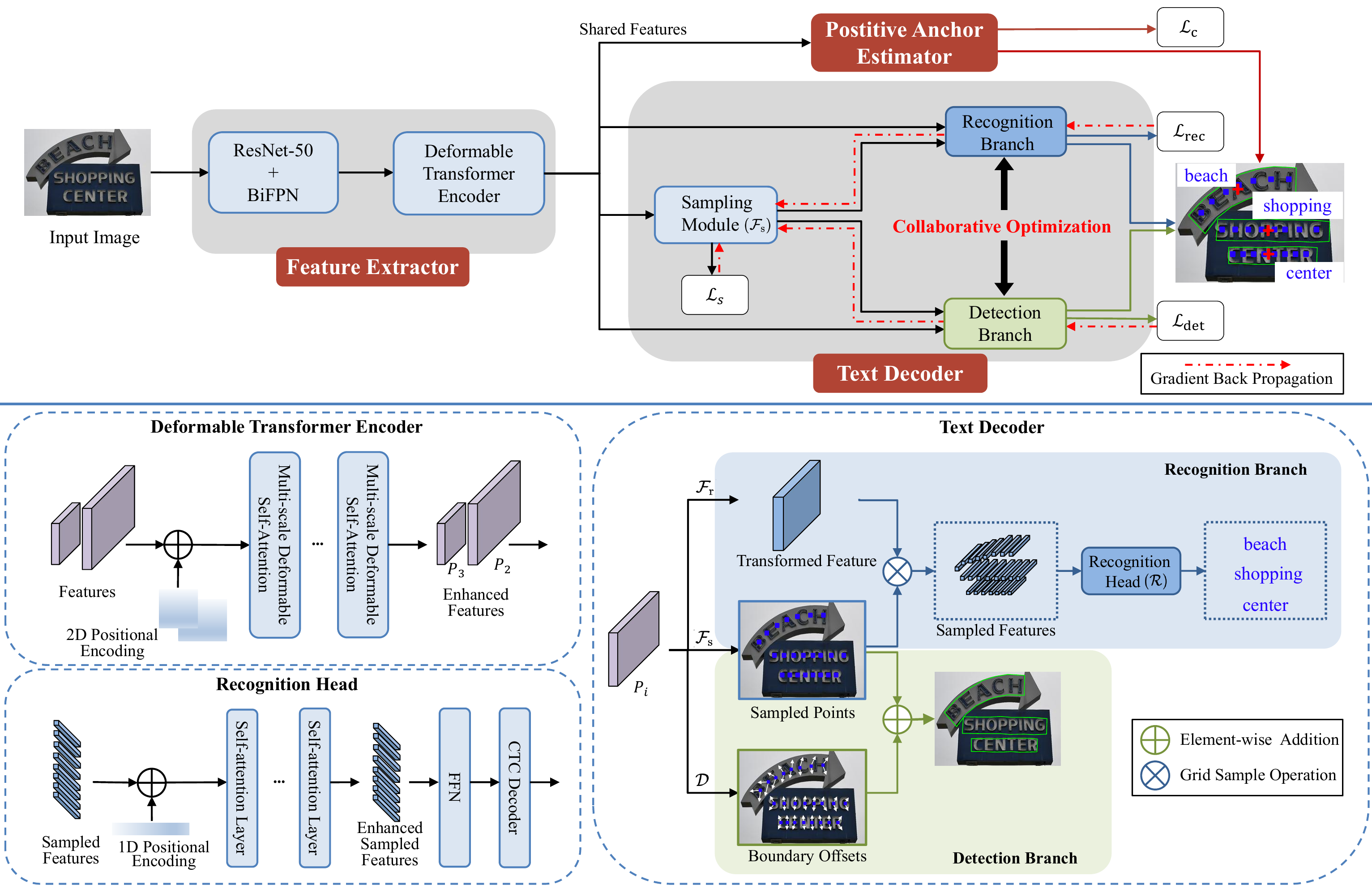}}
  \caption{Architecture of the proposed single shot Self-Reliant Scene Text Spotter v2 (\textbf{SRSTS v2}). It consists of three modules: Feature Extractor, Positive Anchor Estimator, and Text Decoder. Taking extracted multi-scale features from Feature Extractor, Positive Anchor Estimator predicts a positive anchor point for each potential text instance to perceive its rough position. Meanwhile, Text Decoder performs sampling around each pixel with the designed Sampling Module. Then \textbf{SRSTS v2} conducts text detection and recognition based on the sampled representative feature points by Detection Branch and Recognition Branch, respectively. Since both supervision from text detection and recognition guide the learning of Sampling Module, \textbf{SRSTS v2} is able to perform collaborative optimization and mutual enhancement between text detection and recognition.}
  \label{fig:Framework}
\end{figure*}
As illustrated in Figure~\ref{fig:Framework}, our \textbf{SRSTS v2} consists of three modules: Feature Extractor, Positive Anchor Estimator, and Text Decoder. Given a scene text image, \textbf{SRSTS v2} first employs Feature Extractor to learn multi-scale features, and then conducts estimation of positive anchor points and text decoding concurrently. Specifically, Positive Anchor Estimator predicts a positive anchor point for each potential text instance to roughly locate it. Meanwhile, Text Decoder of \textbf{SRSTS v2} employs designed Sampling Module to sample representative feature points around each pixel in the feature map. Finally, for each positive anchor point indicating a potential text instance, \textbf{SRSTS v2} conducts precise text detection and recognition in parallel using Text Decoder based on the sampled points for this anchor point. 
%As illustrated in Figure~\ref{fig:Framework}, our \textbf{SRSTS} consists of four modules: Feature Extractor, Anchor Estimator, Text Detector and Text Recognizer. \lyu{Given a text image, the \textbf{SRSTS} first employs the Feature Extractor to learn multi-scale features, and then conducts positive anchor points estimation, text detection and text recognition at the same time. The Anchor Estimator estimates positive anchor points for potential text instances. In the inference stage, for each obtained positive anchor point indicating a text instance, the Text Detector locates precise text boundary, and the Text Recognizer of \textbf{SRSTS} performs sampling around positive anchor point and then performs text decoding based on the features of the sampling points.}

\smallskip\noindent\textbf{Feature Extractor.} Feature Extractor of the initial version of \textbf{SRSTS}~\cite{wu2022decoupling} adopts the similar network structure to BiFPN~\cite{tan2020efficientdet}, which is composed of a downsampling pathway and an upsampling pathway. The downsampling pathway employs a ResNet-50~\cite{he2016deep} as the feature learning backbone while the upsampling pathway generates multi-scale feature maps by fusing the features from both pathways in the corresponding feature levels.  

As shown in Figure~\ref{fig:Framework}, we further enhance the feature learning capability of Feature Extractor in \textbf{SRSTS v2} by applying the deformable transformer encoder~\cite{zhu2020deformable} upon the convolutional base module, which has been validated its excellent performance by recent spotters~\cite{kittenplon2022towards,zhang2022text}. The deformable transformer encoder employs several multi-scale deformable attention modules to process feature maps generated by CNNs and outputs enhanced features, which are further fed into Positive Anchor Estimator and Text Decoder. To be specific, Feature Extractor extracts two different scales of features, whose size are $\frac{1}{4}\times \frac{1}{4}$ and $\frac{1}{8}\times \frac{1}{8}$ of the input image size, respectively. Larger feature maps have smaller  receptive field and are used for spotting smaller size of text instances. %\pei{draw the multi-scale features in Figure 1?}
%%% put the following text into the description of recognition head
%To capture the long-range dependency among sampled points within each text instance, we also process the sampled features with self-attention layers in the recognition head. The effective feature enhancement components bring significant performance gain.

 %\textbf{SRSTS v2} adopts the feature extractor of SRSTS to produce two different levels of feature maps, whose feature map size are $\frac{1}{4}\times \frac{1}{4}$ and $\frac{1}{8}\times \frac{1}{8}$ of the input image size, respectively. 

\smallskip\noindent\textbf{Positive Anchor Estimator.} %Instead of spotting text serially as the previous methods, our \textbf{SRSTS v2} decouples recognition from detection by conducting text detection and recognition separately with the reference of anchor points. 
Instead of performing recognition based on the detection results like the previous methods, our \textbf{SRSTS v2} performs detection and recognition in parallel with the guidance of anchor points. In particular, we view each pixel in the extracted feature maps as an anchor point for text spotting, from which Positive Anchor Estimator of \textbf{SRSTS v2} is responsible to predict positive anchor points to estimate the rough positions of potential text instances. Thus, a positive anchor point serves as a reference point to indicate the existence of one text instance and guide Text Decoder of \textbf{SRSTS v2} to perform text decoding for this text instance during the inference stage. Note that Positive Anchor Estimator and Text Decoder of \textbf{SRSTS v2} can be optimized in parallel during the training stage since Text Decoder uses the groundtruth of positive anchor points rather than the predicted ones.

\smallskip\noindent\textbf{Text Decoder.} With the location guidance of anchor points, Text Decoder of \textbf{SRSTS v2} performs text detection and text recognition in parallel. To decouple the task of text recognition from detection, Text Decoder first samples representative feature points for each anchor point with designed Sampling Module. Then it conducts text detection and recognition based on the sampled points by Detection Head and Recognition Head, respectively. In particular, Sampling Module is optimized by joint supervision from both the recognition and detection, which enables the collaborative optimization and mutual enhancement between two concurrent tasks.

\subsection{Estimation of Positive Anchor Points}
Positive Anchor Estimator is designed to predict positive anchor points for each potential text instance, which 
serve as location reference for both text detection and recognition during the inference stage. To be specific, for each predicted positive anchor point, \textbf{SRSTS v2} samples representative points around it by Sampling Module and conducts text detection and recognition in parallel for the corresponding text instance based on the features of sampled points. Thus, our \textbf{SRSTS v2} is able to decouple text recognition from the detection results. 

We view all pixels in the feature maps as anchor points and select positive anchors from them. To this end, Positive Anchor Estimator learns a confidence map to quantify the probability of being positive anchors for each pixel of feature maps.

% In inference stage, A single anchor point is identified by removing redundant candidates for each text instance.

% The Anchor points are produced in two core steps by the Anchor Estimator: 1) A confidence map is learned to quantify the probability for each pixel of feature maps to be an anchor point; 2) A single anchor point is identified by removing redundant candidates for each text instance.

\smallskip\noindent\textbf{Learning confidence map for anchor points.}
% \subsubsection{Learning confidence map for anchor points.}
The confidence map is learned from the feature maps produced by Feature Extractor. Each pixel value in the confidence map, which is within $[0,1]$, indicates the probability for this pixel to be a positive anchor point. Formally, given a feature map  $\mathbf{P}_i \in \mathbb{R}^{W\times H\times C}$
containing $C$ channels of features with size $W\times H$, 
 a non-linear transformation is performed by a $3\times 3$ convolutional layer and a $1 \times 1$ convolutional layer with Batch Normalization and ReLU in between. Here $1 \times 1$ convolutional layer is applied to reduce the channel number to 1. Finally, Sigmoid function $\sigma$ is used to project confidence values into $[0,1]$. Mathematically, the confidence map $\mathbf{C}_i \in \mathbb{R}^{W\times H}$ is calculated from the feature map $\mathbf{P}_i$  by:
\begin{equation}
    \mathbf{C_i} = \sigma \Big(\mathcal{F}_{\text{conv}^{1\times 1}}\big(\text{BN\_ReLU}(\mathcal{F}_{\text{conv}^{3\times 3}}(\mathbf{P}_i))\big)\Big),
    \label{eqn:confidence_map}
\end{equation}
where $\mathcal{F}_{\text{conv}^{1\times 1}}$ and $\mathcal{F}_{\text{conv}^{3\times 3}}$ denote the operations of $1 \times 1$ and $3\times 3$ convolutional layers, respectively.

Intuitively, a positive anchor point should be close to the center position of its corresponding text instance. Thus, we optimize the parameters for learning the confidence map (in Equation~\ref{eqn:confidence_map}) in a supervised way with the Dice loss~\cite{milletari2016v}: 
\begin{equation}
    \mathcal{L}_c = 1 - \frac{2|\mathbf{C} \cap \mathbf{C_\text{gt}}|}{|\mathbf{C}|+|\mathbf{C_\text{gt}}|}.
\end{equation}
Herein, $\mathbf{C_\text{gt}}$ is the groundtruth for the confidence map, which is derived from the detection groundtruth of text instances represented by bounding boxes or polygons. Specifically, the pixels of the central region for each text instance in $\mathbf{C_\text{gt}}$ are assigned the value of 1 while other pixels are assigned 0. The Dice loss is used to maximize the overlap between the predicted confidence map $\mathbf{C}$ and the groundtruth $\mathbf{C_\text{gt}}$. 

The groundtruth map $\mathbf{C_\text{gt}}$ is generated in a similar way to EAST~\cite{zhou2017east}. Concretely, for each text instance, we first locate the vertical centerline of its annotated polygon boundary and derive a quadrilateral region defined by four nearest polygon points to the centerline. Then we shrink the region by a ratio, which is tuned as a hyper-parameter, to obtain the central region for $\mathbf{C_\text{gt}}$ whose pixels are assigned the value of 1.

\smallskip\noindent\textbf{Multi-scale confidence maps for capturing multi-scale text instances.} Each of multi-scale feature maps from Feature Extractor produces an individual confidence map and each confidence map has a corresponding groundtruth. We assign text instances to groundtruth maps heuristically, and each text instance (in the input image) only appears once in all groundtruth confidence maps. Specifically, larger text instances are assigned to the groundtruth confidence map with smaller size since the corresponding feature map has a larger receptive field and is favorable for spotting larger size of text instances. As a result, we learn multi-scale confidence maps, each of which is responsible for the appropriate size of text instances. %The size thresholds for the assignment of text instances to groundtruth confidence maps are hyper-parameters tuned on a validation set.

Since the size of the characters is generally determined by the height of the text polygon, we assign text instances to different scales of feature maps according to the height of text instances with pre-defined thresholds, which are tuned on a held-out validation set.

\subsection{Text Decoding}
As illustrated in Figure~\ref{fig:Framework}, Text Decoder of \textbf{SRSTS v2} performs sampling for each pixel of feature maps, and conducts text detection and recognition in parallel for each potential text instance indicated by a predicted positive anchor point, using the sampled feature points around this anchor point.
\subsubsection{Weakly supervised sampling around anchor points}
Sampling Module performs sampling by predicting the two-dimensional coordinates of each sampled point for each anchor point (pixel) of feature maps. Sampling Module consists of three $3\times 3$ convolutional layers and one $1\times 1$ convolutional layer, with Batch Normalization and ReLU function between the layers. %Taken
Taking the feature maps $\mathbf{P}_i \in \mathbb{R}^{W\times H\times C}$ from Feature Extractor as input, Sampling Module samples $K$ points for each anchor point in $\mathbf{P}_i$ by predicting the offsets of sampled points to the anchor point, which are denoted as a tensor $\mathbf{O}_i\in \mathbb{R}^{W\times H\times 2K}$. Thus, the coordinates of the sampled points $\mathbf{S}_i\in \mathbb{R}^{W\times H\times 2K}$ can be obtained by adding $\mathbf{O}_i$ to the coordinates of each anchor point. For instance, the coordinates of the $K$ sampled points for an anchor point at $(w, h)$ in the level-$i$ feature map are specified by the vector $\mathbf{S}_i[w, h]$:
\begin{equation}
\begin{split}
    &\mathbf{O}_{i} = \mathcal{F}_\text{s} (\mathbf{P}_i),\\
    &\mathbf{S}_i[w,h] = \mathbf{O}_{i}[w,h] + [w,h],
\end{split}
\end{equation}
where $\mathcal{F}_\text{s}$ denotes the transformation function of Sampling Module.  As the distribution of sequence lengths varies in different benchmarks, we set $K$ to be 25 for word-level annotated benchmarks and 30 for line-level annotated benchmarks empirically.%As the  sequence lengths in English words are almost not very long, we set $K$ to 25 empirically as ~\cite{shi2016end}.

Intuitively, high-quality sampled points around an anchor point are expected to involve all characters in the text instance indicated by this anchor point. To this end, we conduct supervision on Sampling Module to encourage it to sample uniformly along the horizontal centerline of the text instance:
\begin{equation}
    \mathcal{L}_s = \|\mathbf{S} - \mathbf{S}_\text{gt}\|_1,
    \label{eqn:sampling_loss}
\end{equation}
where $\mathbf{S}_\text{gt}$ denotes the groundtruth, which is the coordinates of uniformly distributed $K$ points along the centerline of the text instance. $\mathbf{S}_\text{gt}$ can be easily calculated based on the groundtruth polygon of this text instance. 

Note that sampling along the center line of the polygon is just one of optional ways, rather than the sole way, to improve the sampling quality. Thus we only perform such supervision during the warm-up training stage on the synthetic data rather than the full training stage to guide the sampling process and speed up the convergence, which is equivalent to weak supervision. Additionally, Sampling Module is also supervised jointly by the detection loss in Equation ~\ref{eqn:det} and the recognition loss in Equation~\ref{eqn:ctc} to enable collaborative optimization between detection and recognition and thereby achieve optimized sampling distributions.

\subsubsection{Concurrent Detection and Recognition for Text Decoding}
The detection module of our \textbf{SRSTS}~\cite{wu2022decoupling} is initially designed based on YOLACT~\cite{bolya2019yolact}. We redesign the detection module in \textbf{SRSTS v2} by formulating the detection task as estimating the offsets of text boundaries to the sampled representative points. As a result, both the text detection and recognition are performed based on Sampling Module, resulting in a more integrated and effective text decoder. Compared to the text decoder of initial \textbf{SRSTS}, a prominent merit of such design is that it enables the collaborative optimization between detection and recognition. This is because the supervisions from both tasks contribute to the optimization of Sampling Module, leading to more effective sampling and better performance for both detection and recognition.

%\wu{In this section, we will introduce the Text Decoder in detail. In our published conference version ~\cite{wu2022decoupling}, we only utilize the sampling module to perform sampling for text recognition and adopt the instance segmentation framework YOLACT ~\cite{bolya2019yolact} to locate text. Based on our observation, the weakly-supervised sampled points in ~\cite{wu2022decoupling} are always near the text center line, which makes it possible to perform text detection based on the sampled points. In this work, the detection is simplified as regressing the boundary from the sampled points. As a result, an information interaction mechanism is built based on the sampling module, resulting in a more straightforward framework and better spotting performance.}

\begin{figure}[t]
  \centering
  %  \vspace{-10pt} 
  \includegraphics[width=\linewidth]{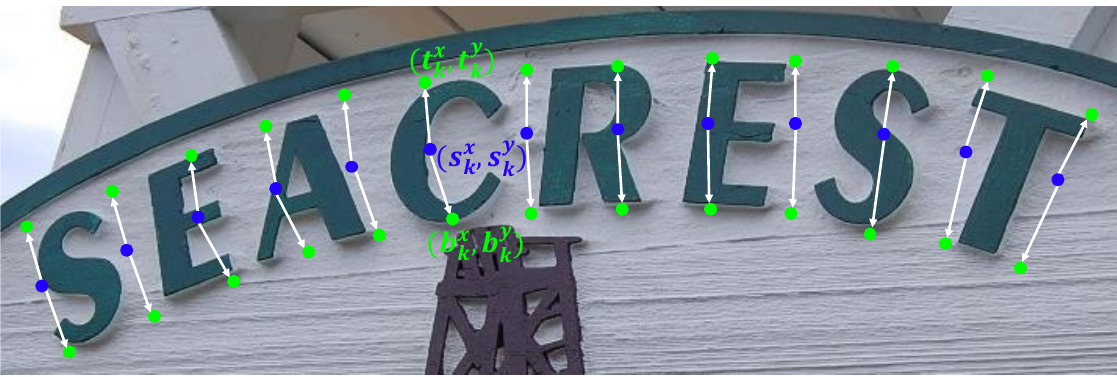}
  \caption{Illustration of text detection by predicting discrete boundary points around the text instance using the sampled points as reference. Referring to a sampled point located at $(s^x_k,s^y_k)$, Detection Head of \textbf{SRSTS v2} correspondingly predicts a boundary point on the top side $(t^x_k,t^y_k)$ and a boundary point at the bottom side $(b^x_k,b^y_k)$, respectively.} 
  \label{fig:detection}
\end{figure}

\smallskip\noindent\textbf{Text Detection.} Sampling Module is guided to sample representative feature points along the centerline of a text instance under the weak supervision indicated in Equation~\ref{eqn:sampling_loss}. The sampled points can be leveraged as reference for text detection, which can potentially lead to more precise detection result. Thus, we design Detection Head $\mathcal{D}$ of our \textbf{SRSTS v2} to conduct detection for each text instance by predicting discrete boundary points around the instance using the sampled points as reference. As shown in Figure~\ref{fig:detection}, Detection Head is supervised to predict two boundary points referring to each sampled point by estimating the offsets of these two boundary points to the sampled point, respectively. The two boundary points are the nearest boundary points to the corresponding sampled point, which locate on the top side and at the bottom side of the boundary, respectively.

To enable parallel computation for efficiency, Detection Head performs detection for each anchor point (pixel) in the feature map, as Sampling Module does, and only the detection results for the positive anchor points (indicating the potential text instances) are used, either for supervision during training or for inference during test. Formally, for each pixel in the feature map $P_i\in \mathbb{R}^{W\times H \times C}$, Detection Head $\mathcal{D}$ predicts $2K$ boundary points corresponding to the $K$ sampled points for this pixel by estimating the offsets between them: 
\begin{equation}
    \mathbf{B}_i = \mathcal{D}(\mathbf{P}_i), %\mathcal{F}_d(\mathbf{P}_i),
\end{equation}
where $\mathbf{B}_i\in \mathbb{R}^{W\times H \times 4K}$ denotes the predicted offsets for two-dimensional  coordinates of $2K$ boundary points for all pixels in the feature map. Thus, the coordinates of the predicted boundary points for the $k$-th sampled point for an anchor point located at $(w,h)$ can be derived by adding the offsets to the corresponding sampled point:
\begin{equation}
\begin{split}
    &\hat{t}_k^x, \hat{t}_k^y = \mathbf{S}_i[w,h, 2k-1:2k] + \mathbf{B}_i[w,h, 4k-3:4k-2], \\
    &\hat{b}_k^x, \hat{b}_k^y = \mathbf{S}_i[w,h, 2k-1:2k] + \mathbf{B}_i[w,h, 4k-1:4k].
\end{split}
\end{equation}
where $(\hat{t}_k^x, \hat{t}_k^y)$ and $(\hat{b}_k^x, \hat{b}_k^y)$ are the coordinates of the predicted boundary points on the top side and at the bottom side of the boundary, respectively. In the training stage, Detection head is supervised by $\text{smooth}_{L_1}$~\cite{girshick2015fast} on the predicted $2K$ boundary points for each of $N$ text instances:
\begin{equation}
\label{eqn:det}
\begin{split}
\mathcal{L}_\text{det}& = \frac{1}{4K}\sum_{n=1}^N \sum_{k=1}^K ( \text{smooth}_{L_1}(\hat{t}_{n,k}^x,t_{n,k}^x)\\
 & + \text{smooth}_{L_1}(\hat{t}_{n,k}^y,t_{n,k}^y)\\
 & + \text{smooth}_{L_1}(\hat{b}_{n,k}^x,b_{n,k}^x)\\
 & + \text{smooth}_{L_1}(\hat{b}_{n,k}^y,b_{n,k}^y)
 ).
\end{split}
\end{equation}
where $(t_{n,k}^y, t_{n,k}^y)$ and $(b_{n,k}^x, b_{n,k}^y)$ are groundtruth of the coordinates of boundary points on the top side and at the bottom side of the boundary corresponding to the $k$-th sampled point for the $n$-th text instance, respectively.

In our implementation, Detection Head $\mathcal{D}$ is constructed as a shallow neural network which consists of three $3\times 3$ convolutional layers and one $1 \times 1$ convolutional layer, with Batch Normalization and ReLU function between the layers. In the inference stage, we perform Non-maximum Suppression (NMS) based on the predicted boundaries and the confidence values of positive anchor points to remove the redundant candidates, and only one positive anchor point is kept for each potential text instance.

\smallskip\noindent\textbf{Text Recognition.}
Our \textbf{SRSTS v2} performs text recognition concurrently with text detection, also relying on the sampled feature points from Sampling Module. Text Decoder of \textbf{SRSTS v2} first transforms the features from Feature Extractor by a recognition transformation module $\mathcal{F}_r$ to adapt to the task of text recognition:
\begin{equation}
    \mathbf{P}^r_i = \mathcal{F}_r (\mathbf{P}_i),
\end{equation}
where $\mathbf{P}^r_i \in \mathbb{R}^{W\times H \times C}$ is the transformed features and $\mathcal{F}_r$ consists of four $3\times 3$ convolutional layers with ReLU and Batch Normalization between the layers. The transformed features are further fed into Recognition Head of Text Decoder to decode the text sequentially. To be specific, for an anchor point located at $(w,h)$, Recognition Head $\mathcal{R}$ takes as input the gathered features $\mathbf{f}_{\pi}\in \mathbb{R}^{K\times C}$ of $K$ sampled points and predicts the sequence of classification probabilities: 
\begin{equation}
\begin{split}
    &\mathbf{f}_{\pi} = \mathbf{P}^r_i[\mathbf{S}^i[w,h]],\\
    &\mathbf{p}_{\pi} = \mathcal{R} (\mathbf{f}_{\pi}),
\end{split}
\end{equation} 
where $\mathbf{p}_{\pi} \in \mathbb{R}^{K\times N_{\text{class}}}$ is the predicted sequence of classification probabilities. $N_{\text{class}}$ denotes the number of character classes which is set according to the specific benchmark. It is set to be 37 for regular word-level English benchmarks (36 for alphanumeric characters and 1 for the blank) while it is set to be 70 for line-level benchmark like CTW1500 (36 for alphanumeric characters, 33 for special symbols, and 1 for the blank). We employ self-attention operation in Recognition Head $\mathcal{R}$ to capture the long-range dependencies among features of sampled points, as shown in Figure~\ref{fig:Framework}. The number of self-attention layer used in $\mathcal{R}$ is tuned as a hyper-parameter in the ablation study (Table~\ref{tab:sa}).

Recognition Head is supervised using the CTC loss ~\cite{graves2006connectionist}:
\begin{equation}
    \mathcal{L}_\text{rec}= \sum^{N}_{n=1} \text{CTC\_loss}(\mathbf{p}_{\pi_n}, \mathbf{q}_{\pi_n}),
    \label{eqn:ctc}
\end{equation}
where $\mathbf{p}_{\pi_n}$ denotes the sequence of the classification probabilities of $K$ characters for the $n$-th text instance while $\mathbf{q}_{\pi_n}$ notates the corresponding groundtruth.

\begin{comment}
is further fed into the recognition head $ \mathcal{R}$.$K$ sampled points are sampled around it to capture the key features for text recognition. The gathered features of $K$ sampled points is further fed into the recognition head $ \mathcal{R}$. The recognition head $ \mathcal{R}$, which consists of several self-attention layers followed by a linear layer, further decodes each text instance based on the feature sequence. The number of self-attention layers is determined according to the results in Table ~\ref{tab:sa}. Formally, for anchor point $n$, $f_{\pi_n}\in \mathbb{R}^{K\times C}$ is the gathered feature sequence of $K$ sampled points. Taking $f_{\pi_n}$ as input, the recognition head $ \mathcal{R}$ outputs the final probability distributions sequence $\mathbf{p}_{\pi_n} \in \mathbb{R}^{K\times N_{\text{class}}}$:
\wu{\begin{equation}
    \mathbf{p}_{\pi_n} = \mathcal{R} (\mathbf{f}_{\pi_n}),
\end{equation} 
$N_{\text{class}}$ denotes the number of class which is set according to the specific benchmark. For regular word-level English benchmarks, $N_{\text{class}}$ is set to 37 (36 for alphanumeric characters and 1 for the blank). As the annotation of CTW1500 is line-level and contains many special symbols, $N_{\text{class}}$ is set to 70 (36 for alphanumeric characters, 33 for symbols, and 1 for the blank).}%For the bilingual task including Chinese and English, the number of the class is set to 5462.   
%Specifically, $K$ boundary points are sampled uniformly at each long side of the text instance.
\end{comment}

\subsubsection{Efficient-Effective Optimization of Text Decoder}
\smallskip\noindent\textbf{Collaborative Optimization between Detection and Recognition.} 
Both text detection and recognition rely on the sampled representative feature points by Sampling Module. Thus, in turn, both the supervision from text detection ($\mathcal{L}_\text{det}$ in Equation~\ref{eqn:det}) and recognition ($\mathcal{L}_\text{rec}$ in Equation~\ref{eqn:ctc}) guide the optimization of Sampling Module in the training stage. As a result, such design enables the collaborative optimization between the text detection and the recognition, which potentially leads to more optimized sampling distributions of feature points and thus yields mutual enhancement between each other. We validate the performance improvement on both text detection and recognition resulting from such collaborative optimization in the experiments in Section~\ref{collaborative_optimization}. 

\smallskip\noindent\textbf{Efficient optimization of Text Decoder by parallel computation.} 
As described previously, in our implementation, all Sampling Module, Detection Head and Recognition Head of Text Decoder are performed for all anchor points (pixels) in the feature map rather than only on the positive anchor points predicted by Positive Anchor Estimator, which allows for the parallel computation between all anchor points. Thus Text Decoder can be optimized quite efficiently and also independently from Positive Anchor Estimator during training. In the inference stage, both Positive Anchor Estimator and Text Decoder are performed in parallel, and only the decoding results (including both detection and recognition) on the positive anchor points are used for the final prediction. Such implementation enables the parallel computation of the whole model and improves the efficiency for both optimization and inference substantially.

\begin{comment}
\subsubsection{Rationale}
\wu{Typical two-stage text spotters depend on text detection to locate the precise text boundaries for text recognition, which implies that the performance of text recognition heavily relies on the detection performance. We argue that the accurate detection of text boundaries is not necessary for text recognition. In our conference version ~\cite{wu2022decoupling}, completely separating text detection and text recognition tasks achieves impressive results. In this paper, our proposed \textbf{SRSTS v2} attempts to increase the interaction between detection and recognition while keeping text recognition independent from precise text detection result. The text recognition and text detection tasks are both performed based on the sampled points, which yields the following advantages for our method: 
1) detection and recognition are conducted simultaneously rather than serially and thus less error propagation from detection to recognition;
2) more information interaction between detection and recognition and both tasks can promote each other by simultaneously guiding the learning of the sampling module;
3) lower annotation cost for text detection.}
\end{comment}

\subsection{End-to-End Parameter Learning}
Following the routine training paradigm adopted by most exiting text spotters~\cite{wang2021pan++,qiao2020mango,liu2021abcnet}, we first pre-train our \textbf{SRSTS v2} on synthetic datasets (`Synthtext'~\cite{gupta2016synthetic} and `Bezier Curve'~\cite{liu2020abcnet}), then train the model on a mixture of synthetic and real-world datasets. Finally, we fine-tune the model on the training set of specific benchmark. The whole model is optimized jointly in an end-to-end manner:
\begin{equation}
\label{eqn:loss}
    \mathcal{L} = \lambda_1\mathcal{L}_\text{c} + \lambda_2\mathcal{L}_\text{s}  + \lambda_3\mathcal{L}_\text{det}+ \lambda_4\mathcal{L}_\text{rec}.
\end{equation}
Herein, $\lambda_1$ to $\lambda_4$ are hyper-parameters to balance between different losses. %To balance different tasks, we set $\lambda_1$, $\lambda_2$, $\lambda_3$ and $\lambda_4$ to 5, 1, 1 and 1 respectively by default. 
Note that $\lambda_2$ is equal to 0 in the joint-training and fine-tuning stage since the sampling supervision $\mathcal{L}_\text{s}$ is only used in the pre-training stage. Besides, the  $\mathcal{L}_\text{s}, \mathcal{L}_\text{det}, \mathcal{L}_\text{rec}$ only work on the positive anchor points.  %To balance the different tasks, we set $\lambda_1$, $\lambda_2$, $\lambda_3$ and $\lambda_4$ to 10, 1, 5, 1 respectively by default.}

\section{Experiment}
%experiment_v1 : ablation studies on tt and ic15
%experiment_v2 : ablation studies on tt and ctw1500

To evaluate the proposed method, we conduct extensive experiments on four popular benchmarks: ICDAR 2015~\cite{karatzas2015icdar}, Rotated ICDAR 2013~\cite{karatzas2013icdar}, CTW1500~\cite{liu2019curved} and Total-Text~\cite{ch2017total}. We first perform ablation studies to investigate the effectiveness of our approach. Then, we compare our model with state-of-the-art methods in terms of performance and inference speed across various challenging spotting scenarios including oriented and curved text appearance. In particular, we make quantitative and qualitative comparisons between our method with ABCNet v2 and TESTR, which are representative methods of two-stage and single shot text spotting paradigms, respectively. %Furthermore, to gain more insight into the proposed sampling-driven spotting strategy on sole text recognition task, we build a standalone recognition model by removing Positive Anchor Estimator and Detection Branch from \textbf{SRSTS v2}, and compare it with a prominent state-of-the-art method for sole text recognition. 

\begin{table*}[!t]
\renewcommand{\arraystretch}{1.1}
  \caption{Comparison between \textbf{SRSTS} and \textbf{SRSTS v2} by ablation study on CTW1500 and Total-Text. `Sampling-based': the text detection is performed based on the sampled points by Sampling Module in \textbf{SRSTS v2}. `DTE': the deformable transformer encoder is employed in Feature Extractor. `SA': self-attention operation is used in Recognition Head of \textbf{SRSTS v2} to capture the long-range dependencies among sampled points. `P’, `R’, `F’ represent `Precision', `Recall' and `F-measure' respectively. `None' and `Full' are two metrics for measuring the end-to-end performance in terms of F-measure. `None’ represents the performance without using lexicon while ‘Full’ corresponds the performance using the lexicon containing all words appearing in the test set.}
  \label{tab:ablative}
  \centering
  \begin{tabular}{l|ccc|ccccc|cccccc}
    \toprule
    %\hline
    \multirow{3}{*}{Method} & \multicolumn{3}{c|}{\multirow{2}{*}{Model Components}} &\multicolumn{5}{c|}{CTW1500} & \multicolumn{5}{c}{Total-Text}\\
    \cmidrule(lr){5-9}
    \cmidrule(lr){10-14}
   % \cmidrule{5-15}
    %\cline{5-15}
    & & & &\multicolumn{3}{c}{Detection} & \multicolumn{2}{c|}{E2E}&\multicolumn{3}{c}{Detection} & \multicolumn{2}{c}{E2E}\\
    \cmidrule(lr){2-4}
    \cmidrule(lr){5-7}
    \cmidrule(lr){8-9}
    \cmidrule(lr){10-12}
    \cmidrule(lr){13-14}
    %\cmidrule{2-14}
   %\cline{2-14} 
     &Sampling-based & DTE & SA& P&R&F& None &Full& P&R&F& None &Full\\
    %\cline{12-13} 
    \midrule
    %\hline
    %SRSTS & & &  & 88.92&83.30 &86.02 &55.59 &78.06 &92.18 & 82.98 &87.34&78.88& 86.50&18.74  \\
    SRSTS~\cite{wu2022decoupling} & & &  & 88.92&83.30 &86.02 &55.59 &78.06 &91.99 & 82.96 &87.24&78.80& 86.33  \\
    \midrule
    %\hline
    SRSTS-sampling  & $\checkmark$ & & & 91.38 & 84.04&87.56& 56.98 & 82.29 & 92.13  &83.97 &87.86&79.75&87.05   \\
    SRSTS-DTE  & $\checkmark$    & $\checkmark$ & &\textbf{91.70}&84.23&87.80&59.57&82.88 &92.42 &86.52 & 89.37 &81.37 & 87.29 \\
    \midrule 
    %\hline
    \textbf{SRSTS v2}  & $\checkmark$   & $\checkmark$ & $\checkmark$  &90.53 & \textbf{86.46}& \textbf{88.45}&\textbf{61.24} & \textbf{83.54 }&\textbf{93.30}& \textbf{86.74}&
\textbf{89.90} &\textbf{82.05}& \textbf{88.05}\\
%\hline
\bottomrule
  \end{tabular}
\end{table*}

\subsection{Experimental Setup}

\smallskip\noindent\textbf{Evaluation benchmarks.}
%We evaluate our model on following challenging benchmarks: ICDAR 2015~\cite{karatzas2015icdar}, Rotated ICDAR 2013~\cite{karatzas2013icdar},  Total-Text~\cite{ch2017total} and CTW1500~\cite{liu2019curved}. 
Four challenging benchmarks are used in the evaluation. 1) ICDAR 2015 contains 1000 training images and 500 testing images. It is annotated with quadrangles and work-level text transcriptions and provides 3 lexicons named ‘Strong’, ‘Weak’, and ‘Generic’ for evaluation. 
2) ICDAR 2013 is a regular text benchmark that contains 229 training images and 223 testing images. In typical experiments~\cite{liao2020mask, huang2022swintextspotter,kittenplon2022towards}, this dataset is rotated at a specific angle for more challenging evaluation.
3) CTW1500 contains 1000 training images and 500 testing images. It is a line-level annotated scene text dataset that contains arbitrary-shaped text instances. A `Full' lexicon is provided which includes all words in the testing set. 
4) Total-Text contains 1255 training images and 300 testing images. It is annotated with polygons and word-level transcriptions. A `Full' lexicon is also provided for evaluation.

\smallskip\noindent\textbf{Datasets for joint training.}
Following the previous methods~\cite{qiao2020mango,liu2021abcnet, zhang2022text,wu2022decoupling}, we combine the following datasets for joint training: Synthtext~\cite{gupta2016synthetic} which is a synthetic dataset containing 800k images; Bezier Curve Synthetic Dataset~\cite{liu2020abcnet} that contains 90k synthetic straight text images and 50k curved text images; COCO-Text~\cite{veit2016coco} which a real-world dataset comprising 63686 images; ICDAR 2017 MLT~\cite{nayef2017icdar2017} that is a multi-language scene text dataset, and ICDAR 2019 ArT~\cite{chng2019icdar2019} containing 5,603 training images. The joint training dataset is a mixture of Synthtext, Bezier Curve Synthetic Dataset, COCO-Text, ICDAR 2017 MLT, ICDAR 2019 ArT, ICDAR 2015, and Total-Text with the sampling ratio 1:1:1:1:2:1:2 correspondingly.

\smallskip\noindent\textbf{Implementation details.}
\label{sec:training}
% The training process is divided into pre-train stage and fine-tune stage. The model is firstly trained with synthetic datasets and then fine-tuned with joint training set. In the pre-train stage, the learning rate is initially set to 2e-3, and decays according to "poly" learning rate strategy. In the fine-tune stage, we set the initial learning rate as 1e-3 and use joint training strategy. The joint training dataset is a mixture of Synthtext, Bezier Curve Synthetic Dataset, COCO-Text, ICDAR 2017(only English samples) , ICDAR 2019 ArT, ICDAR 2015 and TotalText. Our model is optimized by using SGD optimizer with a batch size of 16 on 2 GPUs. 
The training process consists of a warm-up pre-training stage, a joint pre-training stage, and a fine-tuning stage. The model is first optimized in the warm-up pre-training stage on synthetic datasets including Synthtext and Bezier Curve Synthetic Dataset for 300,000 steps of gradient descent, then it is trained with joint training set for another 300,000 steps and finally fine-tuned on the training set of the target benchmark to be evaluated for 30,000 steps.  %4 epochs and then fine-tune the model another 300000 steps with the joint training dataset. The joint training dataset is a mixture of Synthtext, Bezier Curve Synthetic Dataset, COCO-Text, ICDAR 2017 MLT(only English samples) , ICDAR 2019 ArT, ICDAR 2015 and Total-Text. For the fine-tuning task on ICDAR2015 and Total-Text, we set the sample ratio of different datasets to 1:0:1:1:0:1:0 and 1:1:1:1:2:1:2, respectively.  
We use the same data augmentation as \textbf{SRSTS}~\cite{wu2022decoupling} to train our model. In detail, we resize the input image with a randomly selected scale from 0.4 to 1.7 and keep the aspect ratio unchanged.  To handle rotated text well, we also randomly rotate the input image with an angle in the range of [-10$^{\circ}$,10$^{\circ}$]. We randomly crop patches from the input image and resize the longer side to the size of 640 and pad the resized image to $640\times640$ for effective training. In addition, random blur and color jitter are also used. In the inference stage, we resize the longer side of input image to be 1920, 1120, and 704 for ICDAR 2015, Rotated ICDAR 2013, and CTW1500 respectively and the shorter side to be 640 for Total-Text.

We use SGD to optimize our model with the initial learning rate of 1e-3 for the warm-up pre-training stage and the joint pre-training stage. The initial learning rate for the final fine-tuning stage is set to be 1e-5 for the word-level annotated benchmarks and 1e-4 for the line-level annotated benchmark, respectively. We set the weight decay to be 0.0001 and momentum to be 0.9, and delay the learning rate with a `poly’ learning rate strategy~\cite{chen2017deeplab}. Our model is trained with a batch size of 16. %All image samples are resized to be $640 \times 640$ during training. In the inference stage, we resize the longer side of input image to be 1920, 1120, 704 for ICDAR 2015, Rotated ICDAR 2013 and CTW1500 respectively and the shorter side to be 640 for Total-Text.

%tab:encoder
\begin{comment}
\begin{table*}[!t]
\renewcommand{\arraystretch}{1.1}
  \caption{Effect of varying the number of deformable transformer encoder layers in Feature Extractor of \textbf{SRSTS v2}.}
  \label{tab:encoder}
  \centering
  \begin{tabular}{c|ccccc|cccccc}
    \toprule
    \multirow{3}*{\#Layers} &\multicolumn{5}{c|}{CTW1500} & \multicolumn{6}{c}{Total-Text}\\
     %\cline{2-12}
    % \cmidrule{2-12}
    \cmidrule(lr){2-6}
    \cmidrule(lr){7-12}
   &\multicolumn{3}{c}{Detection} & \multicolumn{2}{c|}{E2E} &\multicolumn{3}{c}{Detection} & \multicolumn{2}{c}{E2E} & \multirow{2}*{FPS}\\
   \cmidrule(lr){2-4}
   \cmidrule(lr){5-6}
   \cmidrule(lr){7-9}
    \cmidrule(lr){10-11}
   % \cmidrule{2-11}
   % \cline{12-13} 
   &P&R&F & None & Full&P&R&F & None & Full\\
    \midrule
    %0  & 94.94 & 83.05& 88.60 & 85.31& 81.87   & 77.09& 93.41  &83.55  &88.20&79.40 & 86.37\\
    %0  &94.69 &83.25 & 88.60& 85.47 & 81.63 &76.61 & 93.41  &83.55  &88.20&79.40 & 86.37\\
    0 & 91.38 & 84.04&87.56& 56.98 & 82.29 &
 92.13  &83.97 &87.86&79.75&87.05 & \textbf{20.22}\\
    2 & 91.09&\textbf{84.45}&87.64&56.76&82.28&\textbf{93.97} & 85.20 & 89.37& 80.07&86.76 &17.02\\
    4 & \textbf{91.79}&83.97&87.70&57.31&82.81&92.89 &86.43 & \textbf{89.55} &81.00 & 87.05 &15.22\\ 
    6 &91.70&84.23&\textbf{87.80}&\textbf{59.57}&\textbf{82.88}&92.42 &\textbf{86.52} & 89.37 &\textbf{81.37} & \textbf{87.29} & 13.44\\  
    \bottomrule
  \end{tabular}
\end{table*}
\end{comment}
\begin{table*}[!t]
\renewcommand{\arraystretch}{1.1}
  \caption{Effect of varying the number of deformable transformer encoder layers in Feature Extractor of \textbf{SRSTS v2}.}
  \label{tab:encoder}
  \centering
  \begin{tabular}{c|ccccc|ccccc}
    \toprule
    \multirow{3}*{\#Layers} &\multicolumn{5}{c|}{CTW1500} & \multicolumn{5}{c}{Total-Text}\\
     %\cline{2-12}
    % \cmidrule{2-12}
    \cmidrule(lr){2-6}
    \cmidrule(lr){7-11}
   &\multicolumn{3}{c}{Detection} & \multicolumn{2}{c|}{E2E} &\multicolumn{3}{c}{Detection} & \multicolumn{2}{c}{E2E}\\
   \cmidrule(lr){2-4}
   \cmidrule(lr){5-6}
   \cmidrule(lr){7-9}
    \cmidrule(lr){10-11}
   % \cmidrule{2-11}
   % \cline{12-13} 
   &P&R&F & None & Full&P&R&F & None & Full\\
    \midrule
    %0  & 94.94 & 83.05& 88.60 & 85.31& 81.87   & 77.09& 93.41  &83.55  &88.20&79.40 & 86.37\\
    %0  &94.69 &83.25 & 88.60& 85.47 & 81.63 &76.61 & 93.41  &83.55  &88.20&79.40 & 86.37\\
    0 & 91.38 & 84.04&87.56& 56.98 & 82.29 &
 92.13  &83.97 &87.86&79.75&87.05\\
    2 & 91.09&\textbf{84.45}&87.64&56.76&82.28&\textbf{93.97} & 85.20 & 89.37& 80.07&86.76\\
    4 & \textbf{91.79}&83.97&87.70&57.31&82.81&92.89 &86.43 & \textbf{89.55} &81.00 & 87.05\\ 
    6 &91.70&84.23&\textbf{87.80}&\textbf{59.57}&\textbf{82.88}&92.42 &\textbf{86.52} & 89.37 &\textbf{81.37} & \textbf{87.29}\\  
    \bottomrule
  \end{tabular}
\end{table*}
%tab:sa
\begin{comment}
\begin{table*}[!t]
\renewcommand{\arraystretch}{1.1}
  \caption{Effect of varying the number of the self-attention layers in Recognition Head of \textbf{STSTS v2}.}
  \label{tab:sa}
  \centering
  \newcommand{\tabincell}[2]{\begin{tabular}{@{}#1@{}}#2\end{tabular}}
  %
  %\small
  \begin{tabular}{c|ccccc|cccccc}
    %\hline
    \toprule
    \multirow{3}*{\#Layers} 
     &\multicolumn{5}{c|}{CTW1500} & \multicolumn{6}{c}{Total-Text} \\
    %\cline{2-12}
    \cmidrule(lr){2-6}
    \cmidrule(lr){7-12}
     &\multicolumn{3}{c}{Detection} & \multicolumn{2}{c|}{E2E}&\multicolumn{3}{c}{Detection} & \multicolumn{2}{c}{E2E}&\multirow{2}*{FPS} \\
    \cmidrule(lr){2-4}
    \cmidrule(lr){5-6}
    \cmidrule(lr){7-9}
    \cmidrule(lr){10-11}
   % \cline{12-13} 
   &P&R&F & None & Full &P&R&F & None & Full& \\
    \midrule
    0&\textbf{91.70}&84.23&87.80&59.57&82.88&92.42 &86.52 & 89.37 &81.37 & 87.29 & \textbf{13.44}\\
    2 &91.00&86.16&\textbf{88.52}&60.96&83.20&92.43 & 86.32  &89.27 & 81.81 &87.56 & 13.14
    \\
    4 &90.53 & \textbf{86.46}& 88.45&61.24 & \textbf{83.54 }& 93.30& 86.74&
89.90 &\textbf{82.05}& \textbf{88.05}&12.86\\ 
    6 &91.22&85.08&88.05&\textbf{62.02}&83.16& \textbf{93.46}&\textbf{86.95} & \textbf{90.08}&81.96&87.54 &12.55\\  
    \bottomrule
  \end{tabular}
\end{table*}
\end{comment}

\begin{table*}[!t]
\renewcommand{\arraystretch}{1.1}
  \caption{Effect of varying the number of the self-attention layers in Recognition Head of \textbf{STSTS v2}.}
  \label{tab:sa}
  \centering
  \newcommand{\tabincell}[2]{\begin{tabular}{@{}#1@{}}#2\end{tabular}}
  %
  %\small
  \begin{tabular}{c|ccccc|ccccc}
    %\hline
    \toprule
    \multirow{3}*{\#Layers} 
     &\multicolumn{5}{c|}{CTW1500} & \multicolumn{5}{c}{Total-Text} \\
    %\cline{2-12}
    \cmidrule(lr){2-6}
    \cmidrule(lr){7-11}
     &\multicolumn{3}{c}{Detection} & \multicolumn{2}{c|}{E2E}&\multicolumn{3}{c}{Detection} & \multicolumn{2}{c}{E2E} \\
    \cmidrule(lr){2-4}
    \cmidrule(lr){5-6}
    \cmidrule(lr){7-9}
    \cmidrule(lr){10-11}
   % \cline{12-13} 
   &P&R&F & None & Full &P&R&F & None & Full\\
    \midrule
    0&\textbf{91.70}&84.23&87.80&59.57&82.88&92.42 &86.52 & 89.37 &81.37 & 87.29\\
    2 &91.00&86.16&\textbf{88.52}&60.96&83.20&92.43 & 86.32  &89.27 & 81.81 &87.56
    \\
    4 &90.53 & \textbf{86.46}& 88.45&61.24 & \textbf{83.54 }& 93.30& 86.74&
89.90 &\textbf{82.05}& \textbf{88.05}\\ 
    6 &91.22&85.08&88.05&\textbf{62.02}&83.16& \textbf{93.46}&\textbf{86.95} & \textbf{90.08}&81.96&87.54\\  
    \bottomrule
  \end{tabular}
\end{table*}

\subsection{Ablation Study}

In this section, we conduct ablation studies on CTW1500 and Total-Text, which are line-level annotated and world-level annotated benchmarks respectively, to investigate the effectiveness of core components of \textbf{SRSTS v2}. We perform training with the same training protocol described in Section~\ref{sec:training} for a fair comparison.

\subsubsection{Comparison between \textbf{SRSTS} and \textbf{SRSTS v2}}
\label{sec: comparison}

Compared with the prior version \textbf{SRSTS}~\cite{wu2022decoupling}, \textbf{SRSTS v2} makes two improvements. First, it improves the spotting scheme by redesigning the detection module. Unlike \textbf{SRSTS} employing YOLACT for text detection, \textbf{SRSTS v2} performs both the text detection and recognition based on the sampled representative feature points by Sampling Module, which enables the collaborative optimization between two tasks and thereby allows for mutual enhancement between them. Second, \textbf{SRSTS v2} employs the deformable transformer encoder in Feature Extractor to improve the feature learning and adopt self-attention operation in Recognition Head $\mathcal{R}$ to capture the long-range dependencies among the sampled points. The experimental results of ablation study are presented in Table~\ref{tab:ablative}, which shows the performance gain from each improvement.%improves the spotting scheme by optimizing between the decoupled detection and recognition. The proposed information interaction mechanism is implemented based on the shared sampling module. In addition, we utilize several effective feature enhancement components to obtain richer features for spotting task. The quantitative performance improvement for each extension is shown in Table~\ref{tab:ablative}.}

%\smallskip\noindent\textbf{The effectiveness of proposed spotting scheme.}
The performance gain between \textbf{SRSTS} and \textbf{SRSTS-sampling} shows the superiority of the detection scheme of \textbf{SRSTS v2} over that of \textbf{SRSTS}. \textbf{SRSTS} performs detection with YOLACT by instance segmentation and predicts the bounding box based on the anchor point. In contrast, \textbf{SRSTS v2} predicts the sampled points to bridge the gap between the anchor point and the detection boundaries, which is potentially more accurate than the detection scheme of YOLACT. Moreover, the collaborative optimization between detection and recognition leads to more effective sampling and thus results in better performance in both detection and recognition. For instance, the F-measure of detection increases 1.54\% and 0.62\% on CTW1500 and Total-Text respectively. Meanwhile, the recognition performance is improved by $4.23\%$ in terms of `Full' metric on CTW1500.

%\smallskip\noindent\textbf{The effectiveness of adopted feature enhancement components.}
% anchor vs sampling

%\input{table_4}
%recognition loss
\begin{table}
\renewcommand{\arraystretch}{1.1}
\caption{Effect of the recognition branch on the detection task to validate the advantage of collaborative optimization. `Detection-only' is the ablated variant of \textbf{SRSTS v2} by removing the recognition branch.}  
\label{tab:with_rec}
\centering
\begin{tabular}{l|ccc|ccc}
\toprule
 \multirow{3}*{Method} & \multicolumn{3}{c|}{CTW1500}&\multicolumn{3}{c}{Total-Text}\\
 %\cline{2-7}
 \cmidrule(lr){2-4}
 \cmidrule(lr){5-7}
 &\multicolumn{3}{c|}{Detection}&\multicolumn{3}{c}{Detection}\\
\cmidrule(lr){2-4}
 \cmidrule(lr){5-7}
 & P & R & F & P & R & F \\
\midrule
  Detection-only&88.77  &\textbf{86.76} &87.75 & 90.40 &\textbf{87.44}  &88.89\\ 
 \textbf{SRSTS v2}& \textbf{90.53} & 86.46& \textbf{88.45}&\textbf{93.30}  &86.74&\textbf{89.90} \\ 
\bottomrule
\end{tabular}
\end{table}

We can also observe the performance gain from the enhancement of feature learning by employing the deformable transformer encoder and the modeling of long-range dependencies by self-attention operation in Table~\ref{tab:ablative}, respectively. In particular, the end-to-end F-measure for recognition (`None' metric) is improved by 4.26\% and 2.30\% on CTW1500 and Total-Text respectively. We also study the effect of varying the number of deformable transformer encoder layers and self-attention layers, which are listed in Table~\ref{tab:encoder} and Table~\ref{tab:sa}, respectively. Based on these results, we set the number of deformable transformer encoder layers in Feature Extractor and the number of self-attention layers in Recognition Head $\mathcal{R}$ to be 6 and 4, respectively.

\subsubsection{Effectiveness of Collaborative Optimization}
\label{collaborative_optimization}
%\smallskip\noindent\textbf{Coupling or decoupling of detection and recognition.}
In our \textbf{SRSTS v2}, both text detection and recognition are based on the sampled representative points from Sampling Module. In turn, the Sampling Module is optimized by joint supervision from both detection and recognition, which enables the collaborative optimization between two tasks and thereby yields mutual enhancement. 

The experimental comparisons between `SRSTS-sampling' and `SRSTS' in Table~\ref{tab:ablative} already show that the sampling-based detection strategy in \textbf{SRSTS v2} not only improves the detection performance by a large margin over \textbf{SRSTS} which employs YOLACT for detection, but also boosts the recognition performance substantially. These results reveal that the supervision from the detection branch guides the learning of Sampling Module and potentially yields higher-quality sampling, which can enhance the recognition performance.

\begin{figure*}[!t] 
\renewcommand{\arraystretch}{1.1}
   \centering
   \centerline{\includegraphics[width=0.8\textwidth]{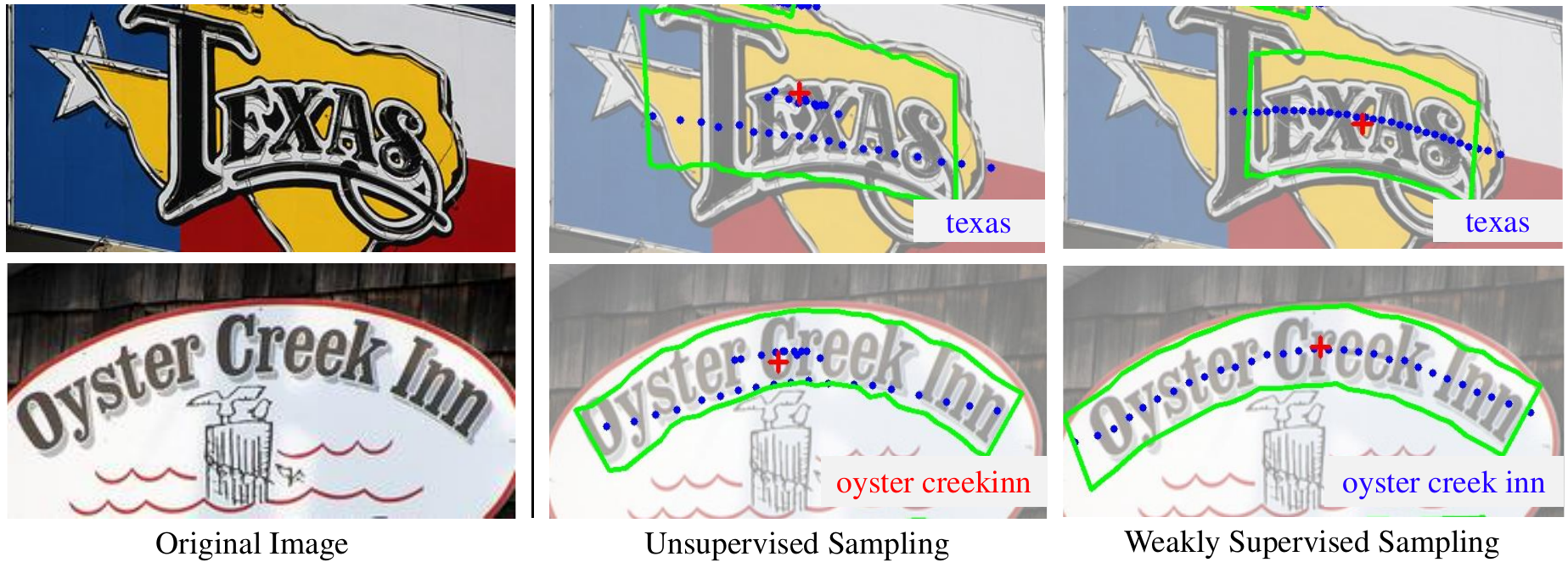}}
  \caption{The visualization of unsupervised sampling and weakly supervised sampling. While the unsupervised sampling can generally produce feasible but somewhat disordered sampled points, weak supervision is sufficient to guide the sampling module to generate a proper distribution of sampling points for detection and recognition along the centerline of the text. The images for showing the results are shaded to visualize the sampled points more clearly.}
  \label{fig:comparison_sampling_mode}
\end{figure*}

%tab:sampling
\begin{table*}[t]
\renewcommand{\arraystretch}{1.1}
  \caption{Comparison among different supervision modes for learning the sampling module of \textbf{SRSTS v2}.}
  \label{tab:sampling}
  \centering
  %\resizebox{\linewidth}{!}{
  \begin{tabular}{l|ccccc|ccccc} 
    \toprule
    \multirow{3}*{Supervision mode}&\multicolumn{5}{c|}{CTW1500}&\multicolumn{5}{c}{Total-Text}\\
   % \cline{2-11}
   \cmidrule(lr){2-6}
   \cmidrule(lr){7-11}
   &\multicolumn{3}{c}{Detection} & \multicolumn{2}{c|}{E2E} &\multicolumn{3}{c}{Detection} & \multicolumn{2}{c}{E2E}\\
   % \cline{2-11}
   \cmidrule(lr){2-4}
   \cmidrule(lr){5-6}
   \cmidrule(lr){7-9}
   \cmidrule(lr){10-11}
   &P&R&F & None & Full&P&R&F & None & Full\\
    \midrule
    Unsupervised sampling &90.20 &85.31 & 87.69& 56.98& 82.75&91.09 &86.37 &88.67 &81.86 &86.92\\ 
    Weakly supervised sampling &90.53 & \textbf{86.46}& \textbf{88.45}&\textbf{61.24} & \textbf{83.54 }&93.30& 86.74&89.90 &\textbf{82.05}& \textbf{88.05} \\ 
    Fully supervised sampling &\textbf{91.46} &85.23 & 88.23&60.08 &82.69 &\textbf{94.19}&  \textbf{86.88}& \textbf{90.39}&81.52&87.54\\ 
    \bottomrule
  \end{tabular}
  %}
\end{table*}

%\limited annotated

\begin{table*}[t]
\renewcommand{\arraystretch}{1.1}
  \caption{Validation of feasibility of learning with limited detection supervision. $\textbf{SRSTS v2}_\text{synthetic}$ is learned with only synthetic data while $\textbf{SRSTS v2}_\text{text}$ is trained with fully-annotated synthetic data and text-only annotated real-world data.}
  \label{tab:weak}
  \centering
  \begin{tabular}{l|ccccc|ccccc}
    \toprule
    \multirow{3}*{Method} &\multicolumn{5}{c|}{CTW1500}&\multicolumn{5}{c}{Total-Text}\\
    \cmidrule(lr){2-6}
    \cmidrule(lr){7-11}
    & \multicolumn{3}{c}{Detection}& \multicolumn{2}{c|}{E2E} &\multicolumn{3}{c}{Detection} & \multicolumn{2}{c}{E2E}\\
    %\cline{5-14}
    \cmidrule(lr){2-4}
    \cmidrule(lr){5-6}
    \cmidrule(lr){7-9}
    \cmidrule(lr){10-11}
   % \cline{12-13} 
     & P& R&F &None&Full& P& R&F& None & Full\\
    \midrule
    $\textbf{SRSTS v2}_\text{synthetic}$ &45.45 &51.64 &48.35 &25.04 & 40.30 &  77.64 &47.70  &  59.09&  43.71& 44.12   \\
    $\textbf{SRSTS v2}_\text{text}$ & \textbf{93.02} &72.36 &81.40&56.08  &76.63  & 83.34 & 64.57 & 72.76 &69.26 & 75.57   \\
    %SRSTS v2$_{\text{point}}$&$\checkmark$ &$\checkmark$   &  &  &  &   &    &  & & &&\\
    \textbf{SRSTS v2} &90.53 & \textbf{86.46}& \textbf{88.45}&\textbf{61.24} & \textbf{83.54 }&\textbf{93.30}& \textbf{86.74}&
\textbf{89.90} &\textbf{82.05}& \textbf{88.05}\\
    \bottomrule
  \end{tabular}
\end{table*}

To further investigate the effect of supervision from the recognition branch on Sampling Module, we conduct another ablation study by removing the recognition branch from \textbf{SRSTS v2} and compare the detection performance between the ablated variant notated  as `Detection-only' and the intact \textbf{SRSTS v2}. As shown in Table~\ref{tab:with_rec}, the detection performance decreases on both two benchmarks, especially Total-Text.%\pei{example}. 
These results demonstrate distinctly the positive effect of the recognition branch on text detection. On the one hand, the recognition results can help filter out the false positives of detection results. On the other hand, the supervision from the recognition branch can guide the optimization of both Sampling Module as well as Feature Extractor, shared by detection and recognition, which leads to better detection results.

Above ablation studies show that the collaborative optimization between the text detection and recognition allows for mutual enhancement between each other, which is a prominent advantage of more integrated spotting framework of \textbf{SRSTS v2} compared with the prior version \textbf{SRSTS}.

\subsubsection{Effect of Different Supervision Modes for Sampling Module}
In this set of experiments, we conduct ablation study to study the effect of different supervision modes ($\mathcal{L}_s$ in Equation~\ref{eqn:sampling_loss}) for Sampling Module of \textbf{SRSTS v2}. To be specific, three supervision modes are performed on Sampling Module respectively for comparison: 1) unsupervised sampling in which $\mathcal{L}_s$ is not applied in any training stage, 2) weakly supervised sampling which performs $\mathcal{L}_s$ only in the warm-up pre-training stage on the synthetic dataset, and 3) fully supervised sampling which performs $\mathcal{L}_s$ throughout all three training stages including warm-up, joint training and fine-tuning stages. 

Table~\ref{tab:sampling} presents the experimental results on both CTW1500 and Total-Text benchmarks. While the unsupervised sampling performs worst among all three modes, its performance for both detection and recognition is not far from that of other two modes with supervision signals, which is indeed encouraging. It implies that the supervision signal for Sampling Module is not critical to the performance, which is reasonable since text detection and recognition only require the sampled points to be distributed uniformly in the text region and involve all characters, but with no strict constraints for the location of sampled points.

%ic15

\begin{table*}[t]
\renewcommand{\arraystretch}{1.1}
  \caption{Quantitative results on ICDAR 2015.  Methods marked with `*' are trained with character-level annotations. `S' (strong) means a customized lexicon of 100 words, including the groundtruth, is given for each image. `W' (weak) implies a lexicon that includes all words that appear in the test set is provided. `G' (generic) denotes a generic lexicon with 90k words.}
  \label{tab:ic15}
  \centering
  \begin{tabular}{l|l|ccccccccc}
    \toprule
   \multirow{2}*{Lexicon}& \multirow{2}*{Model} & \multicolumn{3}{c}{Detection}& \multicolumn{3}{c}{E2E} &\multicolumn{3}{c}{Word Spotting}\\
   % \cline{3-11}
   \cmidrule(lr){3-5}
   \cmidrule(lr){6-8}
   \cmidrule(lr){9-11}
   & &P&R&F &S&W & G &S&W & G\\
    \midrule
    \multirow{13}{*}{\makecell[l]{Official lexicon}}
    &TextNet~\cite{sun2018textnet} &89.42& 85.41&  87.37& 78.66&  74.90& 60.45 &82.38 &  78.43 &  62.36\\
     &FOTS~\cite{liu2018fots} & 91.00&85.17  & 87.99 &  81.09&  75.90&  60.80 & 84.68 & 79.32 & 63.29\\
     &TextDragon~\cite{feng2019textdragon} &  92.45 &83.75 & 87.88 & 82.54&  78.34&  65.15& 86.22&  81.62 & 68.03\\
     &Qin \textit{et al.}~\cite{qin2019towards}  &89.36& 85.75 & 87.52 & 83.38 & 79.94 &  67.98 &- &- &-\\
     &TextPerceptron~\cite{qiao2020text} & 92.30 & 82.50 & 87.10 & 80.50 & 76.60 & 65.10 & 84.10 & 79.40 & 67.90\\
     &PGNet~\cite{wang2021pgnet}  & 91.80 & 84.80& 88.20 & 83.30 & 78.30 & 63.50 & - & - & -\\
     &Boundary~\cite{wang2020all} & 89.80 & \textbf{87.50} & 88.60 & 79.70 & 75.20 & 64.10 & - & - & -\\
     &PAN++~\cite{wang2021pan++} &91.40& 83.90& 87.50 & 82.70 & 78.20 & 69.20& - & - & -  \\
     \cmidrule{2-11}
     &SRSTS (ours)~\cite{wu2022decoupling} &\textbf{96.05}& 81.96 &  88.44 & 82.83  & 80.70  &  69.47 & 87.37  &84.75 & 72.33\\   
     & \textbf{SRSTS v2} (ours) &93.43 &85.60 & \textbf{89.35}& \textbf{83.97}&\textbf{81.16} &\textbf{71.66} &\textbf{88.37} &\textbf{85.20} &\textbf{75.48}\\
    \cmidrule{2-11}
     & CharNet$^{*}$\cite{xing2019convolutional} &91.15 & 88.30 &  89.70 & 80.14&  74.45&  62.18& -&- &-\\
     &CRAFTS$^{*}$~\cite{baek2020character}  &89.00  & 85.30& 87.10&  83.10 & 82.10 &74.90 &- &- & - \\
     &MANGO$^{*}$~\cite{qiao2020mango} & -&  - & - & 81.80 &78.90&  67.30 & 86.40 & 83.10&  70.30  \\
     \midrule
    \multirow{9}{*}{\makecell[l]{Specific lexicon\\ provided by~\cite{liao2019mask}}}
     &Mask Textspotter v2~\cite{liao2019mask}  & 86.60& 87.30& 87.00& 83.00& 77.70& 73.50& 82.40& 78.10& 73.60\\
     &Mask Textspotter v3~\cite{liao2020mask}  & -& -& -& 83.30& 78.10& 74.20&83.10&79.10& 75.10\\
     &MANGO$^{*}$~\cite{qiao2020mango} &-& -& -& 85.40& 80.10& 73.90&85.20&81.10& 74.60\\ 
     &ABCNet v2$^{*}$~\cite{liu2021abcnet} & 90.40&86.00& 88.10& 82.70& 78.50& 73.00&-&-& -\\
     &TESTR~\cite{zhang2022text} & 90.30 & \textbf{89.70} & \textbf{90.00} & 85.20 & 79.40& 73.60 &- & - & -\\
     &SwinTextspotter~\cite{huang2022swintextspotter}&-& -& -& 83.90 &77.30  &70.50&-& -& - \\
     &TTS${}_{\text{poly}}$~\cite{kittenplon2022towards}  &- &- & -& 85.20 &81.70 &77.40 &85.00 &81.50& 77.30\\ &GLASS~\cite{ronen2022glass}&-&- &- & 84.70& 80.10 & 76.30&\textbf{86.80}& 82.50& \textbf{78.80}\\
    \cmidrule{2-11}
     &SRSTS (ours)~\cite{wu2022decoupling}  &\textbf{96.05}& 81.96 &  88.44 & 85.63 & 81.74 &  74.51&  85.84 & 82.61 & 76.82\\ 
    %\cline{2-11}
     &\textbf{SRSTS v2} (ours) &93.43 &85.60 & 89.35& \textbf{86.48}&\textbf{82.41} &\textbf{78.19} &86.72 &\textbf{83.17} &78.73\\
    % &\textbf{SRSTS v2} (ours)  &94.76 &84.40 & 89.28& \textbf{86.42}& \textbf{82.66} & \textbf{78.07}&\textbf{86.70} &\textbf{83.53} &\textbf{78.91} \\
    \bottomrule
  \end{tabular}
\end{table*}

%ic13

\begin{table*}[h]
\renewcommand{\arraystretch}{1.1}
  \caption{Quantitative results on Rotated ICDAR 2013.  Methods marked with `*' are trained with character-level annotations. `P', `R' and `F' denote `Precision', `Recall' and `F-measure' respectively.}
  \label{tab:ic13}
  \centering
  \begin{tabular}{l|cccccc|cccccc}
    \toprule
   \multirow{3}*{Model} & \multicolumn{6}{c|}{Rotation Angle: ${45}^{\circ}$}& \multicolumn{6}{c}{Rotation Angle: ${60}^{\circ}$}\\
   \cmidrule(lr){2-7}
   \cmidrule(lr){8-13}
      & \multicolumn{3}{c}{Detection}& \multicolumn{3}{c|}{E2E} & \multicolumn{3}{c}{Detection}& \multicolumn{3}{c}{E2E}\\
     \cmidrule(lr){2-4}
    \cmidrule(lr){5-7}
    \cmidrule(lr){8-10}
    \cmidrule(lr){11-13}
    &P &R &F &P &R &F &P &R & F &P &R &F \\
    
    \midrule
    CharNet R-50$^*$~\cite{xing2019convolutional} & -&- & 57.20& -& -& 33.90& - & - &58.80 &- &- &9.30\\ 
    Mask Textspotter v2~\cite{liao2019mask} & 64.80 
&59.90 & 62.20 &66.40& 45.80 & 54.20 & 70.50  & 61.20 & 65.50  &68.20  &48.30  &56.60\\ 
    Mask Textspotter v3~\cite{liao2020mask} & 91.60 &77.90  &84.20 &\textbf{88.50} & 66.80  & 76.10  & 90.70   & 79.40  & 84.70   &\textbf{88.50}   &67.60  &76.60\\ 
    SwinTextspotter~\cite{huang2022swintextspotter} & - &-  &- &83.40&72.50 & 77.60& - &-  &-&  84.60 &72.10&77.90 \\ 
    TTS${}_{\text{box}}$~\cite{kittenplon2022towards} & - &-  &\textbf{89.90} &-&- &  80.10 & - &-  & \textbf{89.70}&-& - & 81.00\\  
    TTS${}_{\text{poly}}$~\cite{kittenplon2022towards} & - &-  &88.80 &-&- &  80.40 & - &-  & 87.60&-& - & 80.10\\  
    \midrule
    \textbf{SRSTS v2} (ours) &\textbf{92.68} & \textbf{85.56}&88.97 &85.36&\textbf{78.80}& \textbf{81.95} &\textbf{92.25} &\textbf{86.02}&89.03 &84.31 &\textbf{78.61} &\textbf{81.36} \\
    \bottomrule
  \end{tabular}
\end{table*}

On the other hand, the supervision $\mathcal{L}_s$ is designed to guide Sampling Module to sample uniformly along the centerline of the text instance, which is validated by the performance improvements of both weakly supervised and fully supervised modes over the unsupervised mode. Nevertheless, the performance of these two supervised modes are on par with each other. This is presumably because sampling along the center line of the polygon is just one of optional ways rather than the sole way. Weak supervision is already sufficient to guide the sampling module to generate a proper distribution of points for detection and recognition, as illustrated in Figure~\ref{fig:comparison_sampling_mode}. Thus, we adopt weak supervision for Sampling Module in our \textbf{SRSTS v2}.

\subsubsection{Learning with Limited Detection Supervision}
%\wu{\subsubsection{Training with Limited Boundary Annotations} }
The recognition of our \textbf{SRSTS v2} does not rely on its detection prediction. Benefiting from such decoupling between recognition and detection, \textbf{SRSTS v2} can be potentially trained with only limited detection supervision, which can substantially reduce the annotation cost.
We conduct experiments to investigate the effectiveness of \textbf{SRSTS v2} in such learning setting. Specifically, we provide the recognition supervision on all training data (including the synthetic and real-world datasets) whilst the detection supervision is only performed on the synthetic data whose boundaries can be readily obtained without human annotation. We notate the learned \textbf{SRSTS v2} in such learning setting as $\textbf{SRSTS v2}_\text{text}$. Since the boundary annotation is not provided for the real-world data, $\textbf{SRSTS v2}_\text{text}$ has to match each predicted text instance for a positive anchor point with a text groundtruth by itself for multi-instance images. We use edit distance as the matching metric in our implementation (other matching algorithms are also feasible). Besides, to have a comprehensive comparison, we also evaluate the performance of \textbf{SRSTS v2} learned with only synthetic data denoted as $\textbf{SRSTS v2}_\text{synthetic}$, i.e., the real-world data is not used at all during training.

Table~\ref{tab:weak} presents the comparative performance for both detection and recognition on CTW1500 and Total-Text. We observe that $\textbf{SRSTS v2}_\text{synthetic}$ performs quite poorly for both detection and recognition on two datasets, which implies the large distribution gap between the synthetic data and the real-world data. By contrast, when providing the real-world training data with only recognition supervision, $\textbf{SRSTS v2}_\text{text}$ performs significantly better than $\textbf{SRSTS v2}_\text{synthetic}$ for both detection and recognition. In particular, the recognition performance (`E2E') is even comparable with some classical methods (check Table~\ref{tab:tt} and \ref{tab:ctw}). These impressive results demonstrate the advantage of decoupling recognition from detection: it indeed enables our \textbf{SRSTS v2} to be trained with limited detection supervision. Moreover, the large performance gain for detection from $\textbf{SRSTS v2}_\text{synthetic}$ to $\textbf{SRSTS v2}_\text{text}$ demonstrates the effectiveness of collaborative optimization again that the recognition supervision yields better sampling quality and thereby enhances the detection performance.

\begin{table*}[t]
  \renewcommand{\arraystretch}{1.1}
  \caption{Quantitative results on Total-Text. Methods marked with `*' are trained with character-level annotations. `L' and `S' denote the length of the longer side and shorter side of input images, respectively. \textbf{SRSTS v2-F} is a faster version of \textbf{SRSTS v2} by removing the deformable transformer encoder layers in Feature Extractor and self-attention layers in Recognition Head. }
  \label{tab:tt}
  \centering
  \begin{tabular}{l|l|ccccccc|cc}
    \toprule
    \multirow{2}*{Model}& 
     \multirow{2}*{Scale}&
     \multicolumn{3}{c}{Detection}& \multicolumn{2}{c}{E2E} &\multicolumn{2}{c|}{Word Spotting} &
     \multirow{2}*{\makecell[c]{FPS\\(paper-reported) }}&  \multirow{2}*{\makecell[c]{FPS\\(re-evaluated) }}\\
    \cmidrule(lr){3-5}
    \cmidrule(lr){6-7}
    \cmidrule(lr){8-9}
    
   % \cline{12-13} 
   & &P&R&F &None& Full &None& Full \\
    \midrule
    TextNet~\cite{sun2018textnet}&L: 920&  68.21& 59.45  &63.53 & 54.02 & -&-&-&-&-\\
    TextDragon~\cite{feng2019textdragon}&-& 85.60 &  75.70&  80.30& 48.80 &74.80&-&-&-&-\\
    TextPerceptron~\cite{qiao2020text}&L: 1350&  88.80&81.80& 85.20& 69.70&78.30&-&-&-&- \\
    Boundary~\cite{wang2020all}&L: 1100& 88.90& 85.00&  87.00& 65.00 &76.10&-&-&-\\
    Qin \textit{et al.}~\cite{wang2020all}&S: 600& 83.30 & 83.40&	83.30& 67.80& -&-&-&4.80&-\\
    ABCNet~\cite{liu2020abcnet}&S: 1000&  -& - & -& 63.74& 77.62& 67.10 &81.14&17.90&14.59 \\
    PGNet-A~\cite{wang2021pgnet} &L: 640 & 85.30&\textbf{86.80}& 86.10 &61.70&- &- &-&\textbf{38.20}& 15.37\\
    ABCNet v2~\cite{liu2021abcnet} &S : 1000&  90.20&84.10& 87.00& 67.89& 79.57& 71.82& 83.39&10.00&9.36\\
    PAN++~\cite{wang2021pan++}&S: 512 & 88.40 & 80.50 &84.20 & 64.90 & 75.70 &- &-&29.20 & 15.31\\
    Mask Textspotter v2~\cite{liao2019mask}&S: 1000& 81.80  &75.40  & 78.50& 65.30& 77.40&-&-&-&-\\ 
    Mask Textspotter v3~\cite{liao2020mask} &S: 1000 & -& -& -&71.20& 78.40&-&-&-&-\\
    CharNet H-57$^{*}$~\cite{xing2019convolutional} & -& 88.60&81.00 &84.60&63.60&-&-&-&-\\
    MANGO$^{*}$~\cite{qiao2020mango} &L: 1600 & -& -&-&72.90 &83.60&-&-&4.30&- \\
    CRAFTS$^{*}$~\cite{baek2020character}&L: 1920 &89.50&85.40 &	87.40&78.70 & -&-&- &-&-\\
    %TESTR~\cite{zhang2022text} &L: 1600&  \textbf{93.40} & 81.40&  86.90&-& -& 73.30 & 83.90 &-&-\\
    TESTR~\cite{zhang2022text} &L: 1600&  \textbf{93.40} & 81.40&  86.90 &69.85& 80.51& 73.30 & 83.90 &5.30&8.20\\
    TTS${}_{\text{poly}}$~\cite{kittenplon2022towards}&-& - &-&-& 75.60 & 84.40& 78.20&  86.30&-&- \\ 
    SwinTextspotter~\cite{huang2022swintextspotter}&S: 1000& - &- & 87.20 &- &- & 72.40 &83.00 & -&-\\
    GLASS~\cite{ronen2022glass} &-&-&-&-&76.60& 83.00&79.90& 86.20&-&-\\ 
    \midrule
    SRSTS (ours)~\cite{wu2022decoupling}& S: 640 &91.99&82.96&87.24&78.80&86.33 &81.52 &90.18&18.74&18.74\\
    \textbf{SRSTS v2-F} (ours) & S: 640& 92.13  &83.97 &87.86&79.75& 87.05 &82.66 &90.89 &20.22 &\textbf{20.22}  \\
    \textbf{SRSTS v2} (ours) &S: 640 & 93.30& 86.74&
\textbf{89.90} &\textbf{82.05}& \textbf{88.05}&\textbf{84.66}&\textbf{91.59} & 12.86& 12.86  \\
    \bottomrule
  \end{tabular}
\end{table*}
% ctw1500

\begin{table}[t]
  \caption{Quantitative results on CTW1500. Methods marked with `*' are trained with character-level annotations.}
  \label{tab:ctw}
  \renewcommand{\arraystretch}{1.1}
  \centering
  \begin{tabular}{l|ccccc}
    \toprule
    \multirow{2}*{Model}& 
     \multicolumn{3}{c}{Detection}& \multicolumn{2}{c}{E2E}\\
    \cmidrule(lr){2-4}
    \cmidrule(lr){5-6}
   % \cline{12-13} 
   &P&R&F &None& Full \\
    \midrule
    TextDragon~\cite{feng2019textdragon}&84.50 &82.80 &83.60 &39.70 &72.40\\
    TextPerceptron~\cite{qiao2020text}&87.50 &81.90 &84.60 &57.00 &-\\
    ABCNet~\cite{liu2020abcnet}& -& -& -& 45.20&74.10\\
    ABCNet v2~\cite{liu2021abcnet}& 85.60& 83.80& 84.70& 57.50&77.20\\
    MANGO$^{*}$~\cite{qiao2020mango}& -& -&-&58.90 &78.70 \\
    TESTR~\cite{zhang2022text} & \textbf{92.00}& 82.60& 87.10& 56.00& 81.50\\
    SwinTextspotter~\cite{huang2022swintextspotter}&- & -&88.00 &51.80 &77.00\\
    \midrule
    SRSTS (ours)~\cite{wu2022decoupling}&88.92 &83.30 &86.02 &55.59 &78.06\\
    \textbf{SRSTS v2} (ours)&90.53 & \textbf{86.46}& \textbf{88.45}&\textbf{61.24}&\textbf{83.54}\\
    \bottomrule
  \end{tabular}
\end{table}

%\vspace{5pt}
\subsection{Comparison with State-of-the-art Methods}
In this section, we compare our \textbf{SRSTS v2} with the state-of-the-art methods for text spotting. Specifically, we make two sets of comparisons on two types of benchmarks respectively: oriented text benchmarks and curved text benchmarks. Besides, we also evaluate the efficiency of our model by comparing it with other methods in terms of inference speed. Finally, we particularly compare our model with ABCNet v2 and TESTR, which are representative methods for two-stage and single shot spotting paradigms, respectively.

\subsubsection{Spotting Results on Oriented Text Benchmarks}
\smallskip\noindent\textbf{Results on ICDAR 2015.}
Table~\ref{tab:ic15} presents the detailed comparative results on ICDAR 2015. Considering that the character-level annotations provide much more additional information, we make comparisons between methods only using word-level annotations for training to have a fair comparison and list separately the methods using the character-level annotations for reference. In addition, we also divide the methods into two groups in terms of the lexicon that is used for decoding: the official lexicon and the specific lexicon~\cite{liao2019mask}. To compare comprehensively with other methods, we evaluate our \textbf{SRSTS v2} using each of two lexicons respectively.

As shown in Table~\ref{tab:ic15}, compared with methods trained with word-level annotations, our method achieves the best performance on both detection (except on `R') and recognition ( in terms of both `E2E' and `Word Spotting'). Impressively, our \textbf{SRSTS v2} even outperforms MANGO~\cite{qiao2020mango} and CharNet~\cite{xing2019convolutional} that use character-level annotations for supervision, which shows the effectiveness of our method. When evaluated with the specific lexicon, our method either achieves the best results (in terms of `E2E') or performs on par with the best performance. In particular, \textbf{SRSTS v2} surpasses both recently proposed TESTR~\cite{zhang2022text} and TTS$_{\text{poly}}$~\cite{kittenplon2022towards} by a large margin in terms of F-measure when evaluated with generic lexicon on end-to-end task.

\smallskip\noindent\textbf{Results on Rotated ICDAR 2013.} To evaluate the robustness of text spotters to rotated text, Mask Textspotter v3~\cite{liao2020mask} proposed to rotate the images in ICDAR 2013 in various angles, which is followed by recent methods~\cite{huang2022swintextspotter, kittenplon2022towards}. The spotting results on Rotated ICDAR 2013 are shown in Table~\ref{tab:ic13}. Our method achieves the second place for text detection and the best recognition (E2E) performance in both cases of different rotation angles. Particularly, our method surpasses the state-of-the-art method TTS~\cite{kittenplon2022towards} by $1.55\%$ and $0.36\%$ on Rotation angle ${45}^{\circ}$ and Rotation angle ${60}^{\circ}$ in terms of end-to-end F-measure respectively, which shows the effectiveness and robustness of our method in the challenging scenarios involving rotated text.

\subsubsection{Spotting Results on Curved Text Benchmarks}

\smallskip\noindent\textbf{Results on CTW1500.}
CTW1500 is a challenging benchmark that contains plenty of long text instances with line-level annotations. %Since the proportion of text instances with a length of more than 30 does not exceed 10\% in the training set, we set the number of sampled points $K$ to be 30 on CTW1500. In the inference stage, the longer side of the input image is resized to 704. 
Table~\ref{tab:ctw} shows the experimental results of our \textbf{SRSTS v2} and other methods for text spotting. Our method performs best for both detection and recognition (`E2E'). In particular, it surpasses other methods substantially in terms of both `None' and `Full' for recognition. 

\smallskip\noindent\textbf{Results on Total-Text.}
Total-Text is a popular benchmark that contains various arbitrary-shaped text instances. %In the inference stage, the shorter side of input image is resized to 640 while keeping the aspect ratio unchanged. 
The experimental results on Total-Text are shown in Table~\ref{tab:tt}. %Since some works~\cite{liu2020abcnet,liu2021abcnet,zhang2022text} only report the results on the word spotting task, we report the performance on the word-spotting task as well for comprehensive comparison. Besides, 
%We also use their official codes of to reevaluate their end-to-end performance and inference speed. 
As shown, our method achieves the best performance in both detection and recognition. In particular, the conference version of our method \textbf{SRSTS} already outperforms other methods for recognition while the extended version \textbf{SRSTS v2} further improves the performance by a large margin, outperforming other methods substantially on both `E2E' and `Word Spotting' with (`Full') or without (`None') lexicon. %For instance, \textbf{SRSTS v2} achieves the F-measure of 82.05\% and 88.05\% in terms of `None' and `Full' metrics on end-to-end tasks respectively, outperforming all the previous state-of-the-art methods by a large margin.}

\begin{comment}
\subsubsection{Qualitative Results}
\wu{We visualize the detection results, sampled points, and predicted text transcriptions in Figure~\ref{fig:vis}. As shown, \textbf{SRSTS v2} performs well when facing challenging text instances which are varied in size, orientation, and length. }
\end{comment}

\subsubsection{Inference Speed}
The reported inference speed of different methods may be evaluated in different configurations or settings. To have a fair comparison, we re-evaluate the efficiency of all the methods which release the official codes using the same hardware (1x 3090Ti GPU) in the same experimental setting. Table~\ref{tab:tt} lists both the reported inference speed in their papers and our tested inference speed. Considering that the deformable transformer encoder in Feature Extractor and the self-attention layers in Recognition Head are computationally expensive and account for a large portion of inference time, we develop a faster version of our method by removing the deformable transformer encoder and the self-attention layers to evaluate the inference speed of the core components of our methods. The resulting variant that balances between efficiency and performance is termed as `\textbf{SRSTS v2-F}'. 

\begin{table}[!t]
  \renewcommand{\arraystretch}{1.1}
  \caption{Comparison among ABCNet v2, TESTR, and \textbf{SRSTS v2} in terms of F-measure for detection and recognition. `None' is the F-measure on end-to-end task when performing evaluation without lexicon. `\#Param.' is the number of trainable parameters.}
  \label{tab:comparison}
  \centering
  \begin{tabular}{l|cc|cc|c}
    \toprule
    \multirow{3}*{Method} & \multicolumn{2}{c|}{CTW1500}& \multicolumn{2}{c|}{Total-Text}& \multirow{3}*{\#Param.}\\
    \cmidrule(lr){2-3}
    \cmidrule(lr){4-5}
   &Detection & E2E&Detection & E2E\\
   \cmidrule(lr){2-2}
   \cmidrule(lr){3-3}
   \cmidrule(lr){4-4}
   \cmidrule(lr){5-5}
   &F & None&F & None\\
    \midrule
    ABCNet v2 & 84.70&57.50&87.00 & 67.89& 47.75M\\
    TESTR  & 87.10 & 56.00 &87.10  &69.85  &49.26M\\
    \textbf{SRSTS v2} &\textbf{88.45} & \textbf{61.24}&\textbf{89.90} & \textbf{82.05}& 41.00M\\
    \bottomrule
  \end{tabular}
\end{table}

As shown in Table~\ref{tab:tt}, although the performance of \textbf{SRSTS v2-F} decreases to some degree compared with \textbf{SRSTS v2}, \textbf{SRSTS v2-F} still outperforms other methods by a large margin for both detection and recognition, which demonstrates the effectiveness of the essential model components. More importantly, \textbf{SRSTS v2-F} achieves the fastest inference speed among all methods involved in the comparison, which shows the superiority of our method in terms of efficiency.

%\wu{Since the evaluation environments of different methods are various and the ways of calculating the speed are also diverse, a fair comparison is necessary. We use the official codes and their provided configurations to retest the speed of some real-time text spotters~\cite{liu2020abcnet,wang2021pgnet,wang2021pan++,liu2021abcnet} in the same hardware (3090 Ti GPU) and calculate the time cost from inputting the input image to outputting the final results. As shown in Table~\ref{tab:tt}, when the deformable transformer encoder and the self-attention operation in Recognition Head are removed, our method \textbf{SRSTS v2-F} surpasses all previous methods in terms of accuracy and efficiency. With a larger model size (the model is equipped with the deformable encoder and self-attention layers), \textbf{SRSTS v2} can achieve better performance with a slight degradation in speed.}

\subsubsection{Comparison with Representative Methods}
In this section, we compare our \textbf{SRSTS v2} with ABCNet v2 and TESTR particularly, which are two representative methods of two-stage and single shot methods for text spotting, respectively. We not only compare the general performance for text detection and recognition between these methods on CTW1500 and Total-Text datasets, but also measure the sensitivity of recognition performance to the detection performance for different methods. Finally, we also perform qualitative comparison to obtain more insight into their difference.

%for make a detailed comparison to understand the advantages of the proposed \textbf{SRSTS v2} compared with traditional two-stage methods and recently proposed single shot methods. ABCNet v2 and TESTR which show superiority over other two-stage methods and single shot methods are selected to be the counterparts. We first report the quantitative results in terms of some evaluation metrics, which are the `None' metric for the end-to-end task, the recognition error rate when IoU of detected polygon and groundtruth is greater than 0.5, and the number of trainable parameters. Then, we visualize some end-to-end results of ABCNet v2, TESTR, and our \textbf{SRSTS v2} to intuitively understand the effectiveness of our proposed method. 

\smallskip\noindent\textbf{\textbf{SRSTS v2} vs. ABCNet v2.} ABCNet v2 is a classical two-stage text spotting method, which encodes text boundary as Bezier curve and employs FCOS~\cite{tian2019fcos} to conduct text detection. The generated detection proposals are further fed into the recognition head for recognition. We use the official code\footnote{\label{website}\url{https://git.io/AdelaiDet}} and their provided pretrained model for evaluation. As reported in Table~\ref{tab:comparison}, our \textbf{SRSTS v2} performs distinctly better than ABCNet v2 for both text detection and recognition, meanwhile it has relatively smaller model size, which reflects the advantages of \textbf{SRSTS v2} over ABCNet v2. 

\begin{figure}[!t] 
\renewcommand{\arraystretch}{1.1}
   \centering
   \centerline{\includegraphics[width=1.0\linewidth]{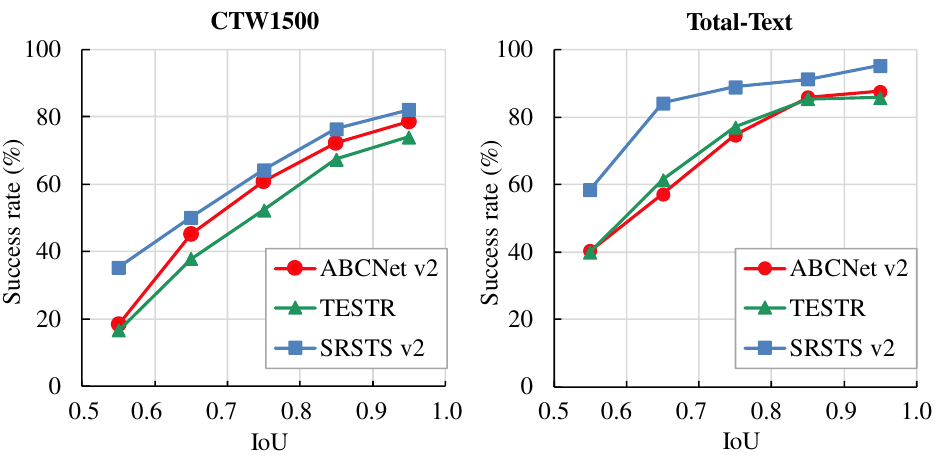}}
  \caption{The success rate of recognition as a function of detection performance to measure the sensitivity of recognition to the detection performance. Our \textbf{SRSTS v2} exhibits larger performance superiority than ABCNet v2 and TESTR at lower detection IoU, which reveals less dependencies between recognition and detection and more robustness of our \textbf{SRSTS v2} than the other two methods.}
  \label{fig:sensitivity}
\end{figure}

\begin{figure*}[!t] 
   \centering  \centerline{\includegraphics[width=\textwidth]{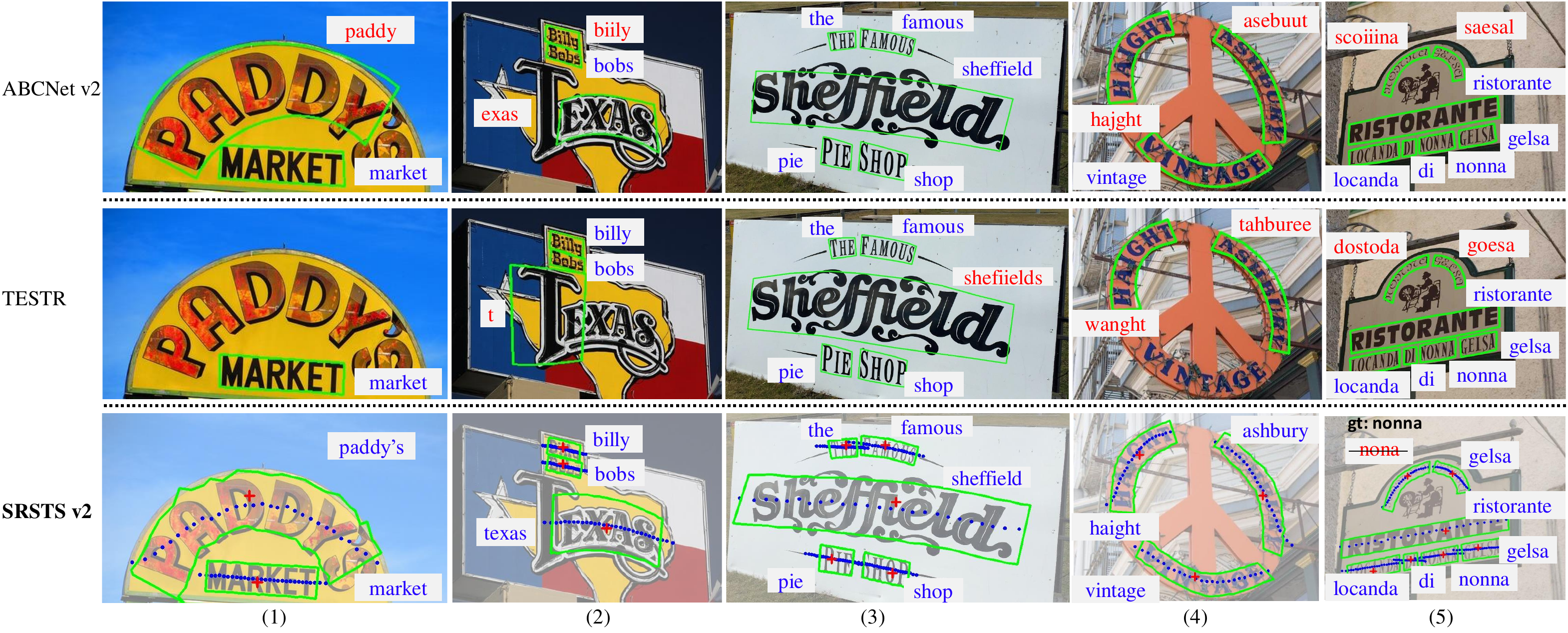}}
  \caption{Visualization of text spotting results of ABCNet v2, TESTR and our \textbf{SRSTS v2} on five challenging cases. The \textcolor[RGB]{255,0,0}{red} ‘$+$’ represents the location of positive anchor point and the \textcolor[RGB]{0,0,255}{blue} dots denote the sampled points. The \textcolor[RGB]{0,255,0}{green} lines show the predicted text boundaries. The images for showing the results of \textbf{SRSTS v2} are shaded to visualize the sampled points more clearly.} 
  \label{fig:vis}
\end{figure*}

A prominent advantage of our \textbf{SRSTS v2} is the decoupling of recognition from detection by conducting both detection and recognition based on the sampling module concurrently, which can reduce the dependencies of text recognition on the detection and circumvent the potential error propagation from detection to recognition. To validate such advantage, we measure the sensitivity of recognition to detection in Figure~\ref{fig:sensitivity} by calculating the success rate of recognition as a function of detection performance measured by IoU. We observe that our \textbf{SRSTS v2} always outperforms ABCNet v2 at different levels of detection performance. More importantly, \textbf{SRSTS v2} exhibits larger performance superiority than ABCNet at lower detection IoU, which reveals less dependencies between recognition and detection and more robustness of our \textbf{SRSTS v2} than ABCNet v2.

%Benefiting from the absence of RoI operations, our \textbf{SRSTS v2} avoids the disadvantages of RoI cropping and the error propagation from text detection. To prove this point, we calculate the proportion of recognition error in the detection results that the IoU of the detected polygon and groundtruth is greater than 0.5. As shown, the error rate of \textbf{SRSTS v2} is considerably lower than that of ABCNet v2.

\smallskip\noindent\textbf{SRSTS v2 vs. TESTR.}
TESTR is a single shot text spotting method that performs both text detection and recognition based on the guidance of the learned query embeddings. We also use their official code\footnote{\label{website}\url{https://github.com/mlpc-ucsd/TESTR.}} for evaluation. As shown in Table~\ref{tab:comparison}, our \textbf{SRSTS v2} outperforms TESTR on both text detection and recognition with smaller model size. 

Theoretically, TESTR can decouple the recognition from detection since the query embeddings for detection and recognition are learned independently. However, figure~\ref{fig:sensitivity} shows that our model has much less dependencies between recognition and detection than TESTR, especially at lower detection performance (indicated by IoU). We surmise that both detection and recognition of TESTR rely heavily on the quality of learned queries. Since TESTR adopts the similar strategy of learning query embeddings to conduct both detection and recognition, the challenging text instances for detection to learn effective queries are also difficult for recognition. In contrast, our \textbf{SRSTS v2} has no strict requirement for the location of the sampled points. It is able to recognize the text instances correctly as long as the sampled points can involve all characters, even though the precise text boundaries are challenging to detect.

%has a larger model size than \textbf{SRSTS v2}, and performs worse in terms of `None' and `Error Rate'. The results demonstrate that our method uses a more lightweight text decoder while still achieving better performance on the end-to-end task compared with TESTR. We argue that the aforementioned performance difference may come from the following reasons: 1) Our method decodes text instances based on the features of sampled points, which are easier to learn compared with the query embedding used in TESTR;  2) The detection and recognition are both conducted based on the sampled representative points by Sampling Module, and in turn both the detection and recognition tasks guide the learning of the sampled module. Such collaborative optimization makes the detection and recognition enhance other other potentially, while the dual decoders of TESTR only shares the reference points to keep weak correlation between two tasks.%The detection and recognition in our \textbf{SRSTS v2} are bridged by shared sampling module, which is more direct and effective than the correlation realized by the shared reference points between the dual decoders of TESTR.}

\smallskip\noindent\textbf{Qualitative Evaluation.} 
We further perform qualitative comparison between our \textbf{SRSTS v2}, ABCNet v2 and TESTR by visualizing the spotting results of three methods on five challenging cases in Figure~\ref{fig:vis}. We make following observations. First, the examples in (1) and (2) show that ABCNet v2 and TESTR fail to recognize the text correctly if the bounding boxes of the text are difficult to be precisely detected whilst our \textbf{SRSTS v2} can still make correct recognition. These examples show the merit of \textbf{SRSTS v2} that decoupling the detection from detection can potentially alleviate the error propagation from detection to recognition. Second, the examples in (3) and (4) illustrate the challenging cases in which ABCNet v2 and TESTR may make wrong recognition due to the complicated text appearance or background even if the text detection is precise. In contrast, our method is able to recognize them correctly, which implies the robustness of our method for recognition. Finally, we also show a failure case in Figure~\ref{fig:vis} (5), in which all three methods fail to make correct prediction for the word `nonna' probably due to the rare typeface that never appears in the training data.

\section{Conclusion}
In this work, we have presented the single shot Self-Reliant Scene Text Spotter v2 (\textbf{SRSTS v2}), which decouples recognition from detection to circumvent the error propagation from detection to recognition. To be specific, the proposed \textbf{SRSTS v2} estimates a positive anchor point for each potential text instance and meanwhile performs sampling for each anchor point by a specially designed sampling module. As a result, \textbf{SRSTS v2} is able to conduct both text detection and recognition in parallel based on the sampled representative feature points, eliminating the dependencies of recognition on the detection results. Moreover, both the supervision from detection and recognition tasks guide the learning of the sampling module, which enables the collaborative optimization and mutual enhancement between detection and recognition. 
%we propose to perform sampling with a sampling module, which is guided by the supervision from the detection and recognition tasks concurrently. As a result, the precise detection of text boundaries, which is demanded by typical two-stage text spotters, is not essential for our \textbf{SRSTS v2}. In addition, the sampling module enables the collaborative optimization between the detection and recognition, which makes the two tasks enhance each other potentially. Besides, the deformable transformer encoder and self-attention operation are employed to obtain richer feature representations.
Benefiting from the proposed sampling-driven concurrent spotting framework, our \textbf{SRSTS v2} outperforms existing methods for text spotting by a large margin. Extensive experiments on four challenging benchmarks demonstrate the effectiveness and advantages of our proposed method.

\ifCLASSOPTIONcaptionsoff
  \newpage
\fi

%\clearpage



%\bibliographystyle{IEEEtran}
\bibliographystyle{IEEEtran}
\bibliography{reference}
%\begin{thebibliography}{1}


%\end{thebibliography}

\begin{IEEEbiography}[{\includegraphics[width=1in,height=1.25in,clip,keepaspectratio]{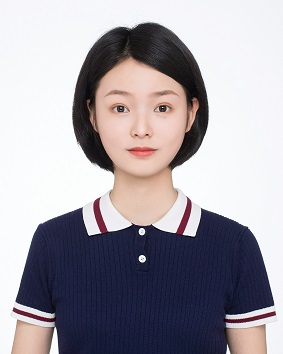}}]{Jingjing Wu}
received her B.S. degree from the School of Computer Science, Wuhan University, Wuhan, China in 2021. She is currently a master student with the Harbin Institute of Technology, Shenzhen, China. Her main research interests include scene text detection and recognition.
\end{IEEEbiography}

\begin{IEEEbiography}[{\includegraphics[width=1in,height=1.25in,clip,keepaspectratio]{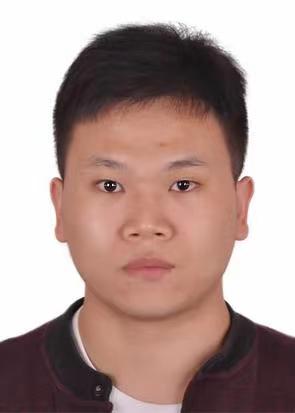}}]{Pengyuan Lyu}
received his B.S. and M.S. degrees from the School of Electronic Information and
Communications, Huazhong University of Science and Technology (HUST), China in 2015 and
2018, respectively. He is currently a senior engineer at Computer Vision Department, Baidu Inc.
His research and development interests mainly focus on scene text detection, recognition, and  document image analysis and understanding, etc.
\end{IEEEbiography}

\begin{IEEEbiography}[{\includegraphics[width=1in,height=1.25in,clip,keepaspectratio]{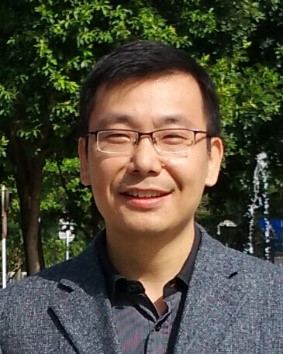}}]{Guangming Lu}
received the undergraduate degree in electrical engineering, the master degree in control theory and control engineering, and the Ph.D. degree in computer science from the Harbin Institute of Technology (HIT), Harbin, China, in 1998, 2000, and 2005, respectively. From 2005 to 2007, he was a Postdoctoral Fellow at Tsinghua University. Now, He is a Professor at Harbin Institute of Technology, Shenzhen, China. He has published over 100 technical papers at some international journals and conferences, including IEEE TIP, TNNLS, TCYB, TCSVT, NeurIPS, CVPR, AAAI, IJCAI, etc. His research interests include computer vision, pattern recognition, and machine learning.
\end{IEEEbiography}

\begin{IEEEbiography}[{\includegraphics[width=1in,height=1.25in,clip,keepaspectratio]{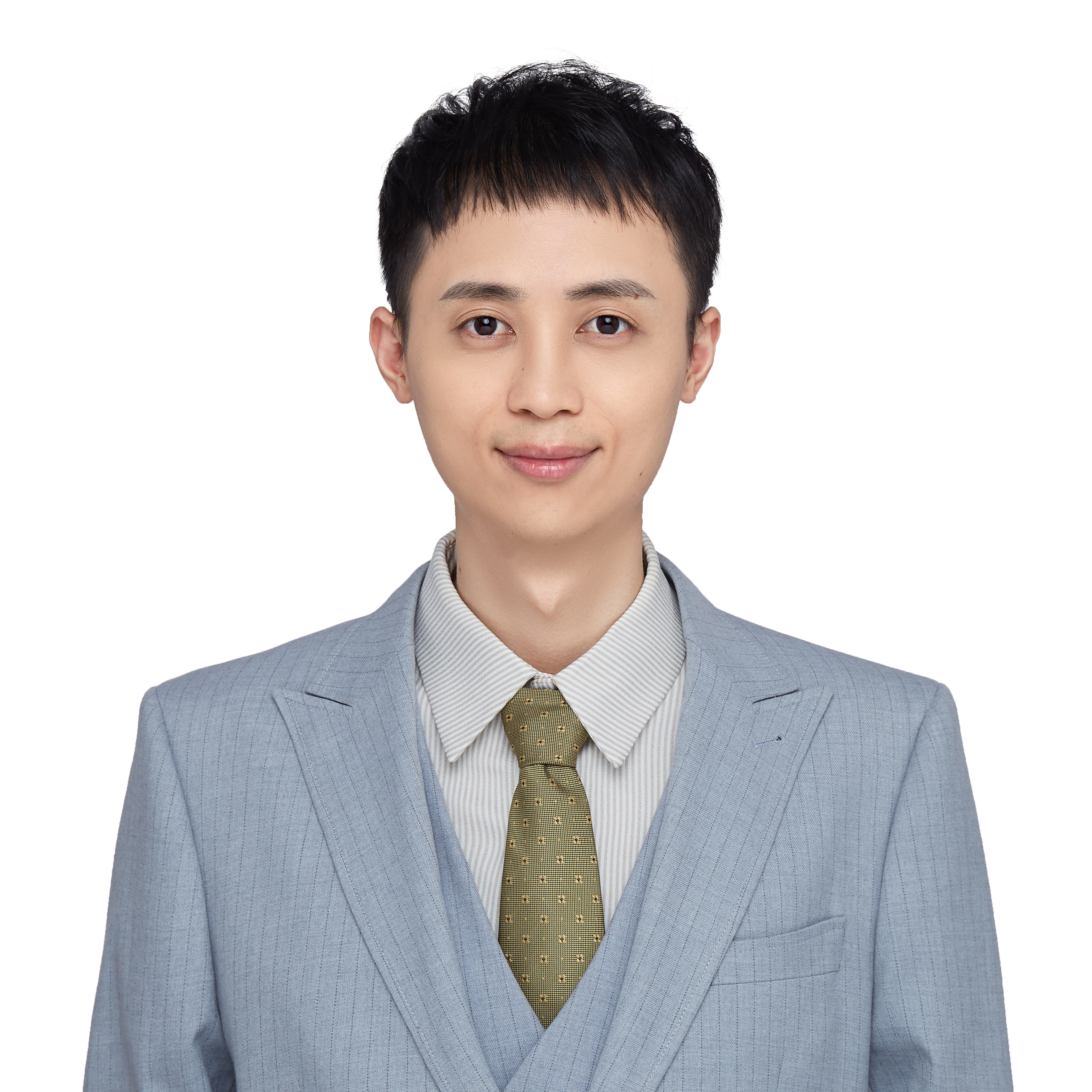}}]{Chengquan Zhang}
received the M.S. degree from Huazhong University of Science
and Technology in 2016. He is currently a Staff Engineer at Computer Vision Department,
Baidu Inc. His research and development interests mainly focus on scene
text detection, recognition and tracking, document image analysis and understanding,
text image editing or generation, etc. 
\end{IEEEbiography}

\begin{IEEEbiography}[{\includegraphics[width=1in,height=1.25in,clip,keepaspectratio]{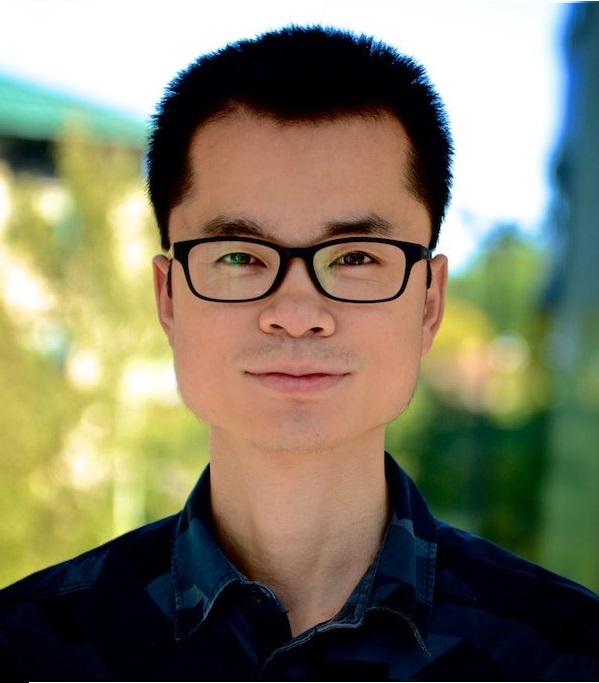}}]{Wenjie Pei} is currently an Associate Professor with the Harbin Institute of Technology, Shenzhen, China. He received the Ph.D. degree from the Delft University of Technology, working with Dr. Laurens van der Maaten and Dr. David Tax. Before joining Harbin Institute of Technology, he was a Senior Researcher on Computer Vision at Tencent Youtu X-Lab. In 2016, he was a visiting scholar with the Carnegie Mellon University. His research interests lie in Computer Vision and Machine Learning.
\end{IEEEbiography}

%\end{IEEEbiography}


%\newpage


%\vfill


\end{document}